\documentclass{article}
\usepackage{ijcai24}

\usepackage{times}
\usepackage{soul}
\usepackage{url}
\usepackage[hidelinks]{hyperref}
\usepackage[utf8]{inputenc}
\usepackage[small]{caption}
\usepackage{graphicx}
\usepackage{amsmath}
\usepackage{amsthm}
\usepackage{booktabs}
\usepackage{algorithm}
\usepackage{algorithmic}
\usepackage[switch]{lineno}

\usepackage{microtype}
\usepackage{subcaption}
\usepackage{booktabs} % for professional tables
\usepackage{multirow}
\usepackage{colortbl}  
\usepackage[svgnames]{xcolor}
\usepackage{array}  
\usepackage{amssymb}

\title{Exploring Learngene via Stage-wise Weight Sharing \\
for Initializing Variable-sized Models}

\author{
Shi-Yu Xia
\and
Wenxuan Zhu\and
Xu Yang\footnotemark[1]\And
Xin Geng\thanks{Co-corresponding author.}\\
\affiliations
School of Computer Science and Engineering, Southeast University, Nanjing 210096, China\\
Key Laboratory of New Generation Artificial Intelligence Technology and Its Interdisciplinary Applications (Southeast University), Ministry of Education, China\\
\emails
\{shiyu\_xia, zhuwx, xuyang\_palm, xgeng\}@seu.edu.cn
}
\begin{document}

\maketitle

\begin{abstract}
In practice, we usually need to build variable-sized models adapting for diverse resource constraints in different application scenarios, where weight initialization is an important step prior to training.
The Learngene framework, introduced recently, firstly learns one compact part termed as \textbf{learngene} from a large well-trained model, after which learngene is expanded to initialize variable-sized models.
In this paper, we start from analysing the importance of guidance for the expansion of well-trained learngene layers, inspiring the design of a simple but highly effective Learngene approach termed SWS (\textit{\textbf{S}tage-wise \textbf{W}eight \textbf{S}haring}), where both learngene layers and their learning process critically contribute to providing knowledge and guidance for initializing models at varying scales.
Specifically, to learn learngene layers, we build an auxiliary model comprising multiple stages where the layer weights in each stage are shared, after which we train it through distillation.
Subsequently, we expand these learngene layers containing \textit{stage information} at their corresponding stage to initialize models of variable depths.
Extensive experiments on ImageNet-1K demonstrate that SWS achieves consistent better performance compared to many models trained from scratch, while reducing around \textbf{6.6$\times$} total training costs.
In some cases, SWS performs better only after \textbf{1 epoch} tuning.
When initializing variable-sized models adapting for different resource constraints, SWS achieves better results while reducing around \textbf{20$\times$} parameters stored to initialize these models and around \textbf{10$\times$} pre-training costs, in contrast to the pre-training and fine-tuning approach.
\end{abstract}

%%%%%%%%%%%%%%%%%%%%%%%%%%% Introduction
\section{Introduction}
\label{intro}
Vision Transformers (ViTs) have become increasingly popular, showcasing their remarkable performance across a wide range of vision tasks~\cite{dosovitskiy2020image,liu2021swin,wang2022image,oquab2023dinov2}.
In practical deployment, it is often necessary to train models of \emph{various scales} to flexibly accommodate different resource constraints.
These constraints may exhibit significant diversity, such as mobile devices with limited available resources and computing centers with substantial computational capabilities.
Clearly, training each target model from scratch provides a straightforward solution, where weight initialization is a crucial step prior to training which aids in model convergence and affects the final quality of the trained model~\cite{glorot2010understanding,he2015delving,arpit2019initialize,huang2020improving}.

Nowadays, a variety of large-scale pretrained models, developed by the research and industry community, are readily available to transfer and finetune the learned weights for diverse downstream tasks~\cite{radford2021learning,he2022masked,touvron2023llama,oquab2023dinov2}.
However, such scheme needs to reuse the original whole pretrained model parameters every time facing different downstream tasks regardless of the available resources.
Unfortunately, for many pretrained model families (MAE~\cite{he2022masked}), even the smallest model (86M ViT-Base~\cite{dosovitskiy2020image}) can be considered extremely large for some resource-constrained settings.
To tackle this, developers would have to first pre-train target model to meet certain resource demand, which is time-consuming, computationally expensive and lacks the flexibility to initialize models of \emph{varying scales}.
Thus, how to flexibly initialize diverse models to satisfy different resource constraints arises as an important research question.

\begin{figure*}[ht]
    \centering
        \includegraphics[scale=0.32]{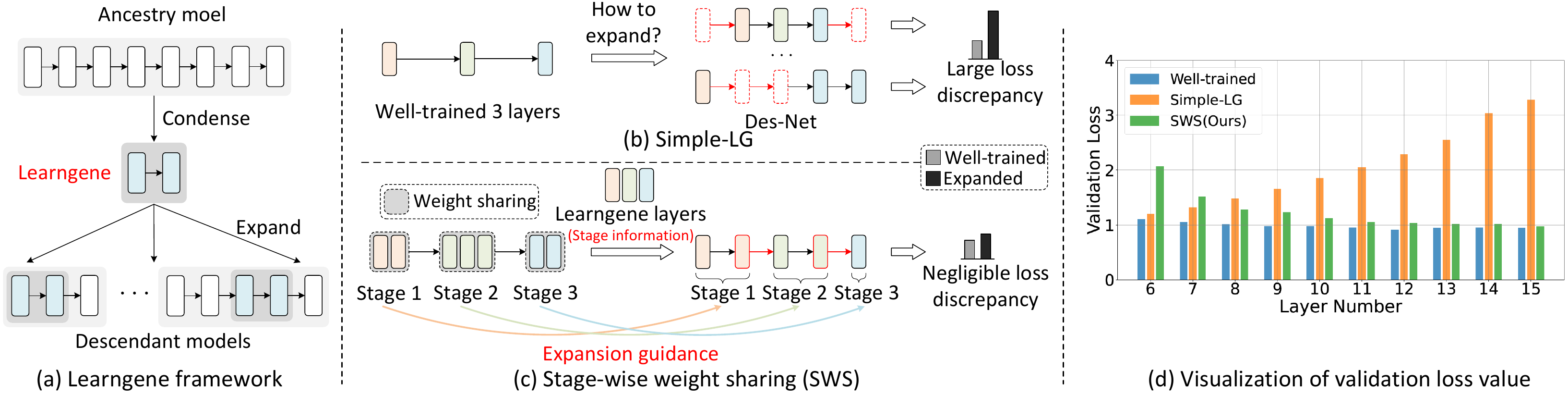}
        % \fbox{\rule{0pt}{2.5in} \rule{.95\linewidth}{0pt}}
    \caption{
    (a) Learngene. (b) Simple-LG and (c) SWS, here we take 3-layer learngenes as an example. 
    (d) Visualization of validation loss value.
    }
    \label{fig:intro2}
\end{figure*}

Recently,~\cite{wang2023learngene} proposes a novel learning paradigm known as \emph{Learngene} to achieve this goal. 
As shown in Fig.~\ref{fig:intro2}(a), Learngene firstly learns one compact part termed as \textbf{learngene}, which contains generalizable knowledge, from a large well-trained network termed as ancestry model (Ans-Net).
Subsequently, learngene is expanded to initialize variable-sized descendant models (Des-Net), after which they undergo normal fine-tuning.
Based on the gradient information of Ans-Net, Vanilla-LG~\cite{wang2022learngene} extracts a few high-level layers as learngene and combines them with randomly-initialized layers to build Des-Nets.
TLEG~\cite{xia2024transformer} extracts two layers as learngene which is linearly expanded to build varying Des-Nets. 
LearngenePool~\cite{shi2024building} distills one large model into multiple small ones whose layers are used as learngene instances and then stitches them to build Des-Nets.
However, there exist several limitations in previous works.
Firstly, the extracted learngene itself only contains well-trained parameters but lacks crucial knowledge essential for the subsequent expansion process.
Secondly, the learngene learning process provides insufficient guidance about how to effectively and conveniently expand the learngene, thereby constraining its potential for initialization. 
Thirdly, empirical performance still lags far behind the pre-training and fine-tuning approach.

To validate the importance of such knowledge and guidance, we study from an idea termed Simple-LG: we first train a vanilla 5-layer model whose layers compose learngene layers, and then expand its layer to initialize Des-Nets of required depth, as shown in Fig.~\ref{fig:intro2}(b).
% Please see more details in the appendix.
Obviously, neither the well-trained layers nor the learning process of the 5-layer model possess knowledge or guidance for the subsequent expansion process.
As a result, this strategy brings severe performance degradation compared to well-trained ones.
For example, it leads to a 27.1\%  accuracy degradation of initialized 12-layer Des-Net in ImageNet-1K~\cite{deng2009imagenet} classification without any fine-tuning.
Moreover, we display the validation loss value of these initialized models in Fig.~\ref{fig:intro2}(d), which delivers two important messages:
1) These initialized models are not fine-tuned, yet they attain meaningful loss value, which demonstrates the potential of expanding learngene layers to initialize models.
2) With the increasing layer number of models, \textit{i}.\textit{e}., the increasing number of expanding layers, the discrepancy between the validation loss of initialized models and that of well-trained ones becomes larger.
For example, expanding the layers to initialize a 12-layer Des-Net (86M) leads to a 150\% relative increase in the validation loss, compared to 47\% observed in a 8-layer one (58M).

Upon closer observation, if we treat each learngene layer as an individual stage, we can \textit{recursively update} each layer at its stage, which equals to sharing weights across multiple layers within each stage throughout the training, shown in left of Fig.~\ref{fig:intro2}(c).
In this way, we seamlessly integrate \textit{stage information} into each learned learngene layer, \textit{i}.\textit{e}., learngene layer actually contains knowledge of multiple weight-shared layers within its stage.
Moreover, we could emulate the learngene expansion process via adding weight-shared layers within each stage during learngene training, thereby providing clear guidance on how to expand, namely, expanding the weight-shared layers at its corresponding stage, shown in Fig.~\ref{fig:intro2}(c).
Both stage information and expansion guidance are necessary: 1) expanding learngene layers which lacks stage information destroys the intrinsic layer connection (See Simple-LG). 2) Without expansion guidance, the position of expanded layers remains uncertain.
Based on this insight, we present \textit{\textbf{S}tage-wise \textbf{W}eight \textbf{S}haring} (\textbf{SWS}), a simple but effective Learngene approach for efficient model initialization.
SWS divides one Transformer into multiple stages and shares the layer weights within each stage during the training process.
Specifically, we design and train an auxiliary model (Aux-Net) whose layer weights are shared via SWS to obtain learngene layers.
Then we can expand these layers at their corresponding stage to initialize variable-sized Des-Nets.
As shown in Fig.~\ref{fig:intro2}(d), we observe that the validation loss of models initialized by SWS can be significantly reduced with the increasing layer numbers.

We systematically investigate the design of sharing stage and the strategy of initialization with learngene.
With extensive experiments, we show the superiority of SWS:
1) Compared to training from scratch, SWS achieves better performance with much less training efforts on ImageNet-1K.
Take Des-B as an example, SWS performs better while reducing around \textbf{6.6$\times$} total training costs.
2) When transferring to downstream classification datasets, SWS surpasses existing Learngene methods by a large margin, \textit{e}.\textit{g}., \textbf{+5.6\% }on Cars-196.
3) When directly evaluating on ImageNet-1K without any tuning after initialization, SWS outperforms existing initialization methods by a large margin, \textit{e}.\textit{g}., \textbf{+9.4\%} with Des-B (86M).
4) When building variable-sized models, SWS achieves better results while reducing around \textbf{20$\times$} parameters stored to initialize and around \textbf{10$\times$} pre-training costs, in contrast to the pre-training and fine-tuning approach.
Our main \textbf{contributions} are summarized as follows:
\begin{itemize}
\item We propose a simple but effective Learngene approach termed SWS for efficient model initialization, which is the first work to systematically explore the potential of weight sharing for initializing variable-sized models.
\item We present the design of weight sharing within each stage and the strategy of initialization with learngene, which firstly highlights the importance of stage information and expansion guidance.
\item Extensive experiments demonstrate the effectiveness and efficiency of SWS, \textit{e}.\textit{g}., compared to training from scratch, training with compact learngenes can achieve better performance while reducing huge training costs.
\end{itemize}

\begin{figure*}[ht]
    \centering
        \includegraphics[scale=0.43]{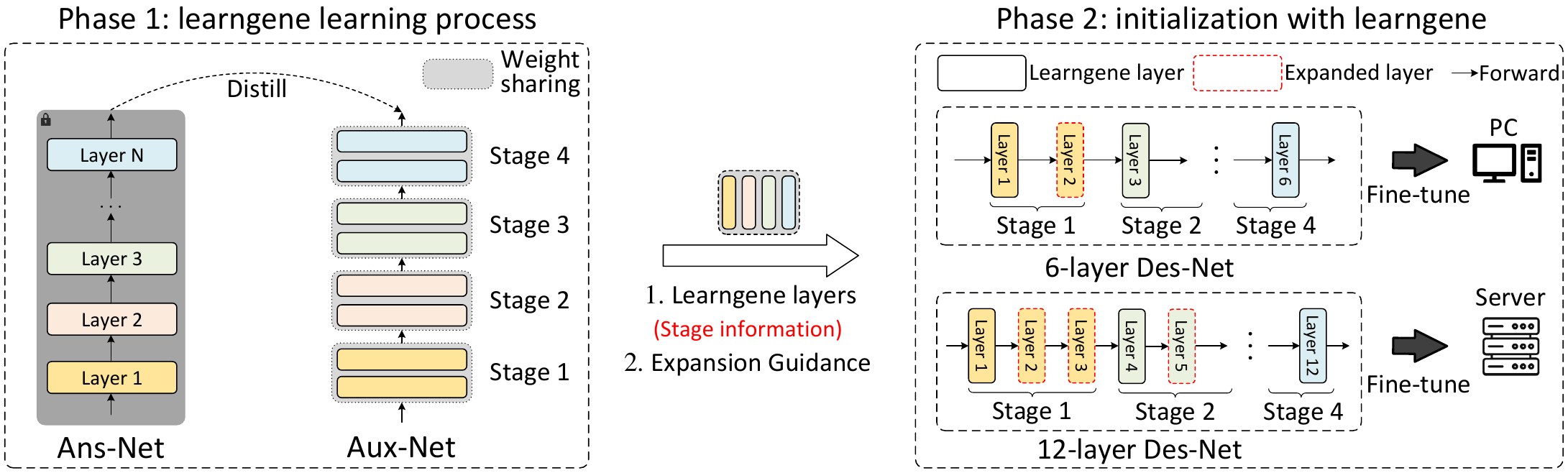}
    \caption{
    In the first phase, we build an auxiliary model comprising multiple stages.
    The layer weights in each stage are shared.
    Note that the number of layers in each stage and the number of stages are both configurable.
    Then we train it via distillation.
    After the learngene learning process, learngene layers containing stage information and expansion guidance are adopted to initialize descendant models of variable depths in the second phase.
    Finally, these models are fine-tuned normally and deployed to practical scenarios with diverse resource constraints.
    }
    \label{fig:method}
\end{figure*}

%%%%%%%%%%%%%%%%%%%%%%%%%%% Related Work 
\section{Related Work}
\subsection{Weight Initialization}
\label{RW_WI}
Weight initialization is a pivotal step prior to training one model and crucially affects the model performance~\cite{glorot2010understanding,nair2010rectified,he2015delving,mishkin2015all,arpit2019initialize,huang2020improving}.
Proper initialization aids in model convergence and training efficiency~\cite{lecun2002efficient}, while arbitrary initialization may hinder the training process~\cite{mishkin2015all}.
Comprehensive initialization strategies have been proposed, such as default initialization from Timm library~\cite{paszke2019pytorch}, Xavier initialization~\cite{glorot2010understanding} and Kaiming initialization~\cite{he2015delving}.
Nowadays, a plethora of pretrained models are readily accessible, offering an excellent initialization for fine-tuning models across a range of downstream tasks~\cite{radford2021learning,bao2021beit,he2022masked,oquab2023dinov2}.
However, this approach needs to reuse the \emph{entire model} for each distinct downstream task, irrespective of the available resources.
Furthermore, we need to pre-train again in instances where a pretrained model of the required size is unavailable, which is extremely time-consuming and computationally expensive.
Recently, \cite{xu2023initializing,samragh2023weight} propose to initialize small models with a larger pretrained model.
By contrast, we seek to train compact learngenes \textit{once} via stage-wise weight sharing and then we can initialize variable-sized models.

\subsection{Learngene}
\label{RW_LG}
Learngene proposes to firstly learn a compact part, referred to as learngene, from a large well-trained model termed as ancestry model (Ans-Net)~\cite{wang2023learngene,feng2024transferring}.
Subsequently, learngene is expanded to initialize variable-sized descendant models (Des-Net), after which they undergo normal fine-tuning.
Vanilla-LG~\cite{wang2022learngene} extracts a few high-level layers as learngene and combines them with randomly initialized layers to build Des-Nets.
TLEG~\cite{xia2024transformer} linearly expands learngene which consists of two layers to initialize Des-Nets of varying scales.
LearngenePool~\cite{shi2024building} distills one pretrained model into multiple small ones whose layers are used as learngene instances, after which they are stitched to build Des-Nets.
In contrast, we firstly investigate integration of stage information into learned learngenes and explore obtaining useful guidance about how to expand learngenes from the learngene learning process, thus better initializing Des-Nets.

\subsection{Weight Sharing}
\label{RW_WS}
Weight sharing is a parameter-efficient model compression strategy~\cite{dabre2019recurrent,lan2019albert,takase2021lessons,zhang2022minivit}, which effectively alleviates over-parameterization problem~\cite{bai2019deep,kovaleva2019revealing} in large pretrained Transformers~\cite{devlin2018bert}.
Different from existing works, we seek to initialize variable-sized models using learngene learned via stage-wise weight sharing, which to our knowledge remains unexplored in the literature.

\begin{figure*}[ht]
    \centering
        \includegraphics[scale=0.41]{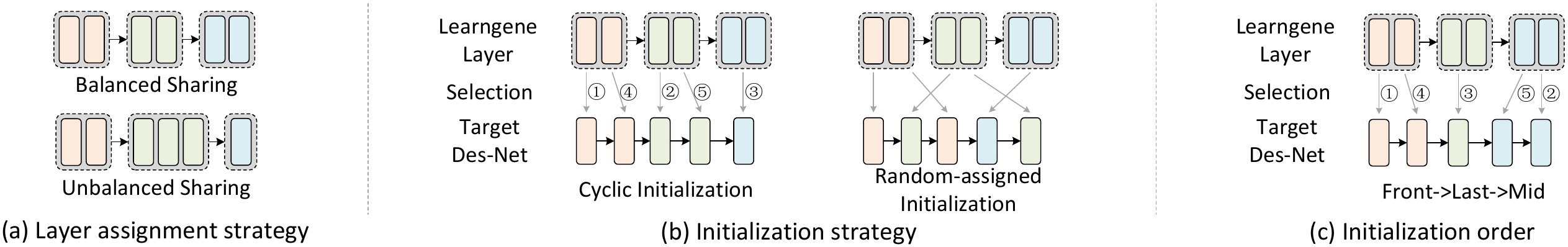}
        % \fbox{\rule{0pt}{2.5in} \rule{.95\linewidth}{0pt}}
    \caption{Taking $M$ = 3 as an example, we show (a) Layer assignment strategy, (b) Initialization strategy and (c) Initialization order.
    }
    % \vspace{-4pt}
    \label{fig:method_2}
\end{figure*}

%%%%%%%%%%%%%%%%%%%%%%%%%%%%%%%%%%%% Approach 
\section{Approach}
\label{approach}
Fig.~\ref{fig:method} depicts the overall pipeline of SWS.
In phase 1, we design an auxiliary Transformer model (Aux-Net) comprising several distinct stages, where the layer weights in each stage are shared.
We train the Aux-Net through distilling from the ancestry model (Ans-Net) to help learn learngene layers and note that the weights of only one layer is trained in each stage due to the sharing mechanism.
In phase 2, the well-trained learngene layers containing \textit{stage information} as well as \textit{expansion guidance} are adopted to initialize descendant models (Des-Net) of variable depths.
Finally, these Des-Nets are fine-tuned normally without the restriction of stage-wise weight sharing.
Next, we firstly introduce some preliminaries.

\subsection{Preliminaries}
Thanks to the modular design of modern vision transformer (ViT)~\cite{dosovitskiy2020image,touvron2021training}, a typical ViT of $L$ layers equipped with parameters $\theta$ can be defined as a composition of functions: $f_{\theta}=f_{L} \circ\cdots\circ f_{1}$, where $f_{\theta}:\mathcal{X} \rightarrow \mathcal{Y}$ transforms the inputs in an input space $\mathcal{X}$ to the output space $\mathcal{Y}$, $f_{i}$ means the function of the $i$-th layer and $\circ$ indicates the composition.
Each layer contains Multi-head Self-Attention (MSA) and Multi-Layer Perceptron (MLP) block, where Layer Normalization (LN)~\cite{ba2016layer} and residual connections are used before and after each block.
The basic idea of weight sharing involves sharing parameters across layers, which is a simple but effective strategy to improve parameter efficiency.
For multi-stage weight sharing, weight sharing in the $m$-th stage can be defined as a recursive update of one shared layer:
\begin{equation}
\label{eq_2}
    Z_{i+1} = H_{m}(Z_{i}, \theta_{m}) , \quad i = 0,1,...,L_{m}-1,
\end{equation}
where $H_{m}$ denotes the function of $m$-th stage, $L_{m}$ denotes the number of weight-shared layers in the $m$-th stage, $Z_{i}$ denotes the representations in the $i$-th layer of $m$-th stage and $\theta_{m}$ represents the parameters of $H_{m}$, \textit{i}.\textit{e}., shared weights of the $L_{m}$ layers.
Note that the parameters of layers in each stage are shared, \textit{i}.\textit{e}., the number of updated parameters in each stage during training equals to that of one layer, but the parameters between stages are not shared.
Thus ViT with parameters $\theta$ can be further defined as $f_{\theta}=H_{M} \circ\cdots\circ H_{1}$ in multi-stage weight sharing setting, where $M$ denotes the number of stages.

\subsection{Stage-wise Weight Sharing of Learngene}
As discussed before, we could seamlessly integrate \textit{stage information} into each learned learngene layer via sharing weights across multiple layers within each stage throughout the training.
Drawing from this insight, we propose to share the parameters of one learngene layer to form multiple weight-shared layers in each stage.
Thus, the number of learngene layers equals to that of stages.
By configuring the number of stages $M$, we can obtain different number of learngenes $\theta_{lg}=\{\theta_{1},...,\theta_{M}\}$.
To ensure clarity, taking the 16-layer ViT-B (114M)~\cite{dosovitskiy2020image} and $M=5$ as an example, $\theta_{lg}$ comprises about 36M parameters which is approximately equivalent to the parameter numbers of five layers and only 36M parameters are updated during the learngene learning process.
In the following, we detail the layer assignment strategy and learngene learning process.

\noindent\textbf{Layer Assignment Strategy.}
Given the number of stages $M$, there are two options to assign the number of layers within each stage: \textit{Balanced} and \textit{Unbalanced} sharing mechanism.
As shown in Fig.~\ref{fig:method_2}(a), Balanced sharing ensures uniformity in the layer numbers across most stages, whereas Unbalanced sharing sets uneven layer numbers within most stages.
However, as Balanced sharing is more aligned with the existing weight sharing principle~\cite{lan2019albert,takase2021lessons}, we will show in Section~\ref{exp_abl} that it achieves more stable and better learngenes than Unbalanced sharing.
In this case, we take Balanced sharing as the default sharing mechanism in SWS.

\noindent\textbf{Learngene Learning Process.}
As the learngene layer is the Transformer layer, but some other components like the patch projection and task-specific head are also required to compose a complete Transformer model.
Therefore, we also add them to build the Aux-Net, after which we train it through distillation. 
Specifically, we consider adopting prediction-based distillation~\cite{hinton2015distilling} to condense knowledge from the Ans-Net, which is achieved by minimizing cross-entropy loss between the probability distributions over the output predictions of the Ans-Net and those of the Aux-Net.
Overall, one distillation loss is defined as:
\begin{equation}
\label{eq_3}
    \mathcal{L}_{d} = CE(\phi(p_{s}/\tau), \phi(p_{t}/\tau)),
\end{equation}
where $CE(\cdot,\cdot)$ denotes soft cross-entropy loss, $p_{t}$ denotes the output logits of the pretrained Ans-Net (\textit{e}.\textit{g}., Levit-384~\cite{graham2021levit}), $p_{s}$ denotes the output logits of the Aux-Net, $\tau$ denotes the temperature value of distillation and $\phi$ denotes the softmax function.
Moreover, we can seamlessly incorporate advanced distillation techniques~\cite{zhang2022minivit,ren2023tinymim,ji2023teachers,li2024accelerating,li2025kdcrowd} into our training process.
Besides distillation, we also introduce one classification loss:
\begin{equation}
\label{eq_4}
    \mathcal{L}_{cls} = CE(\phi(p_{s}), y),
\end{equation}
where $y$ denotes ground-truth label.
Therefore, our total training loss is defined as:
\begin{equation}
\label{eq_5}
    \mathcal{L} = (1-\alpha)\mathcal{L}_{cls} + \alpha\mathcal{L}_{d},
\end{equation}
where $\alpha$ denotes the trade-off.
Noteworthy, the weight sharing constraint always exists during training, \textit{i}.\textit{e}., although Aux-Net contains $L$ layers, only $\theta_{lg}=\{\theta_{1},...,\theta_{M}\}$ which contains parameters of $M$ layers are updated.

\begin{figure*}[ht]
    \centering
        \begin{subfigure}{0.195\linewidth}
        \includegraphics[scale=0.145]{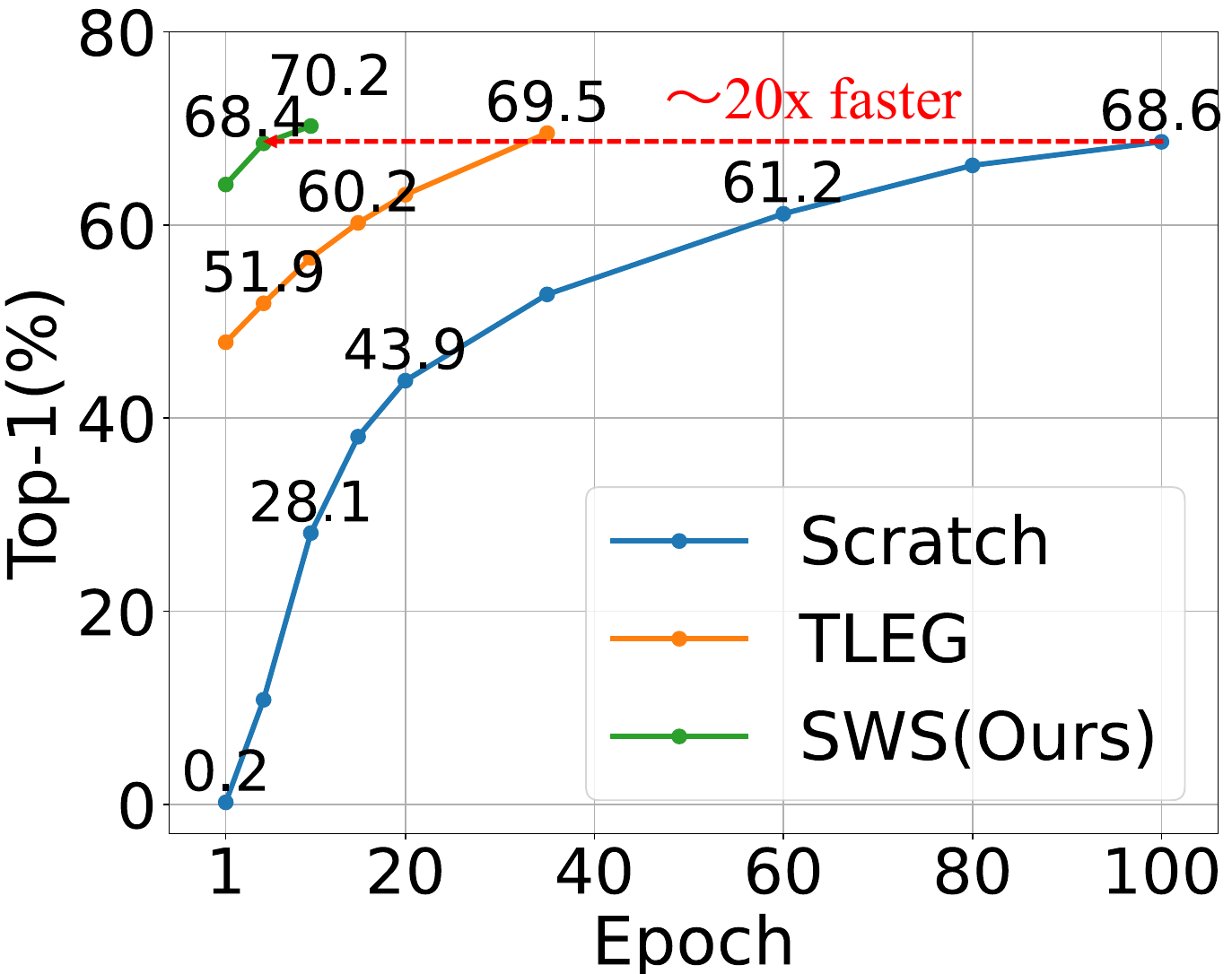}
        % \fbox{\rule{0pt}{1.1in} \rule{.9\linewidth}{0pt}}
        \caption{Des-S-6 (11.4M)}
        \label{fig5_1}
        \end{subfigure}
        \hfill
        \begin{subfigure}{0.195\linewidth}
        \includegraphics[scale=0.145]{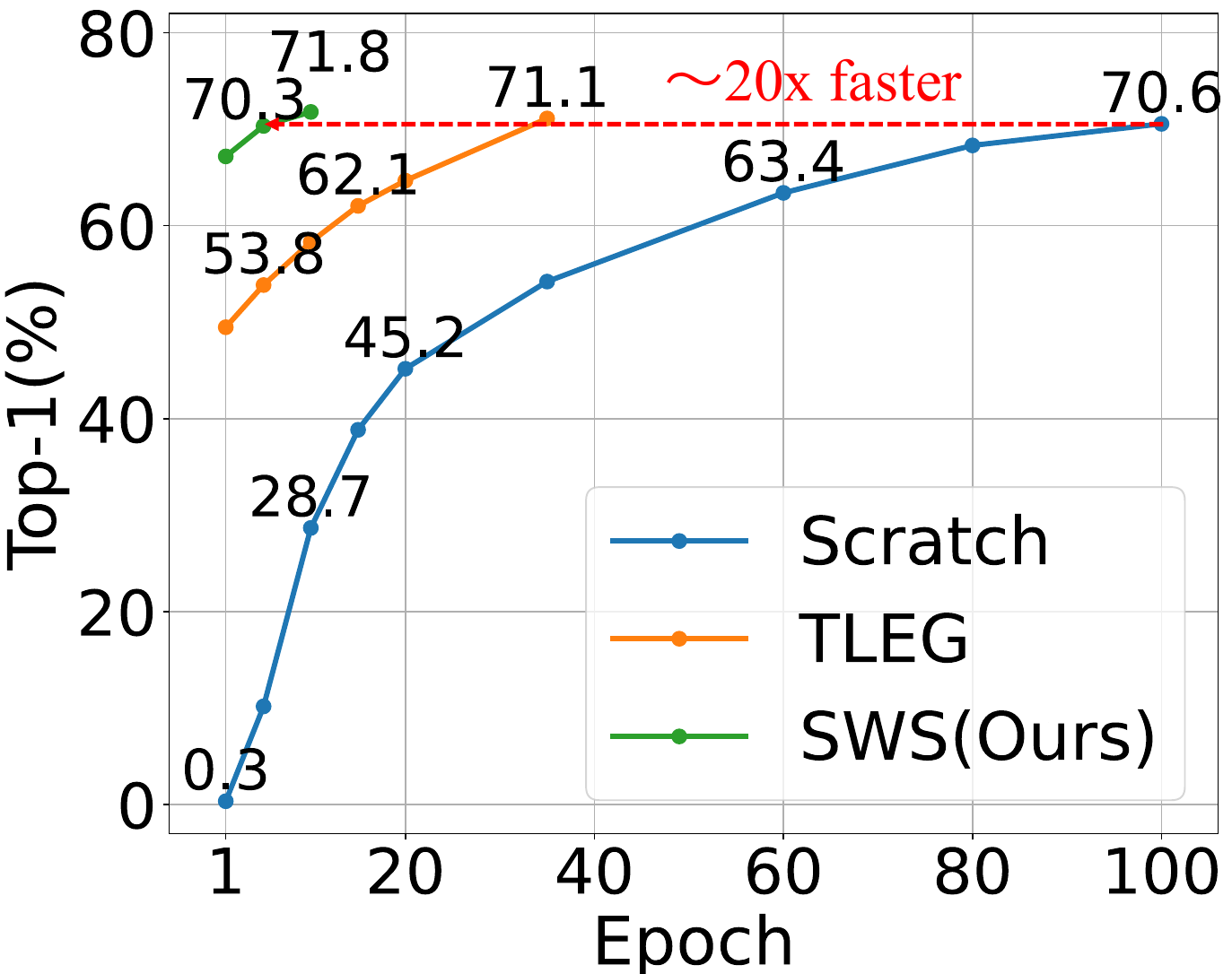}
        % \fbox{\rule{0pt}{1.1in} \rule{.9\linewidth}{0pt}}
        \caption{Des-S-7 (13.2M)}
        \label{fig5_2}
        \end{subfigure}
        \hfill
        \begin{subfigure}{0.195\linewidth}
        \includegraphics[scale=0.145]{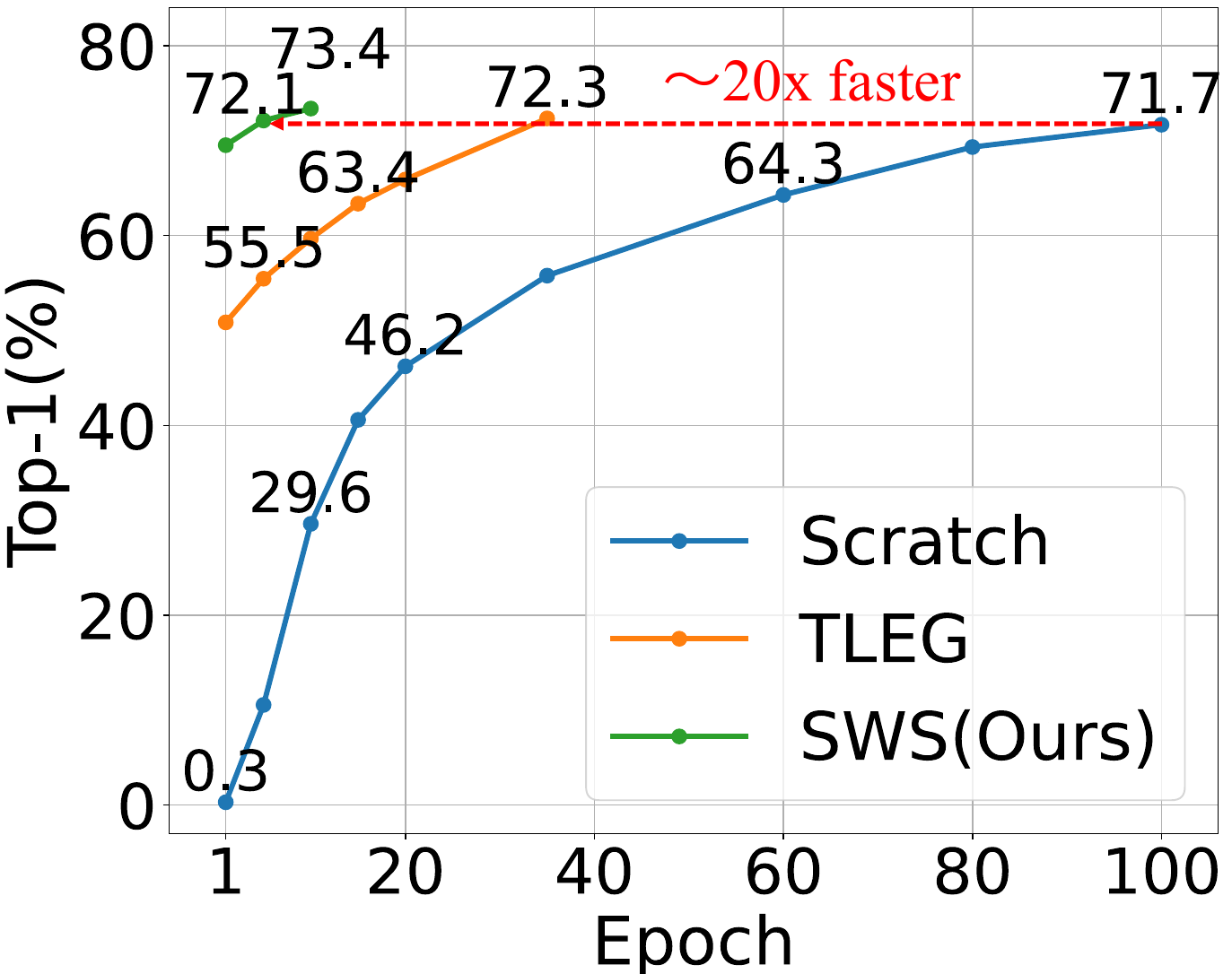}
        % \fbox{\rule{0pt}{1.1in} \rule{.9\linewidth}{0pt}}
        \caption{Des-S-8 (15.0M)}
        \label{fig5_3}
        \end{subfigure}
        \hfill
        \begin{subfigure}{0.195\linewidth}
        \includegraphics[scale=0.145]{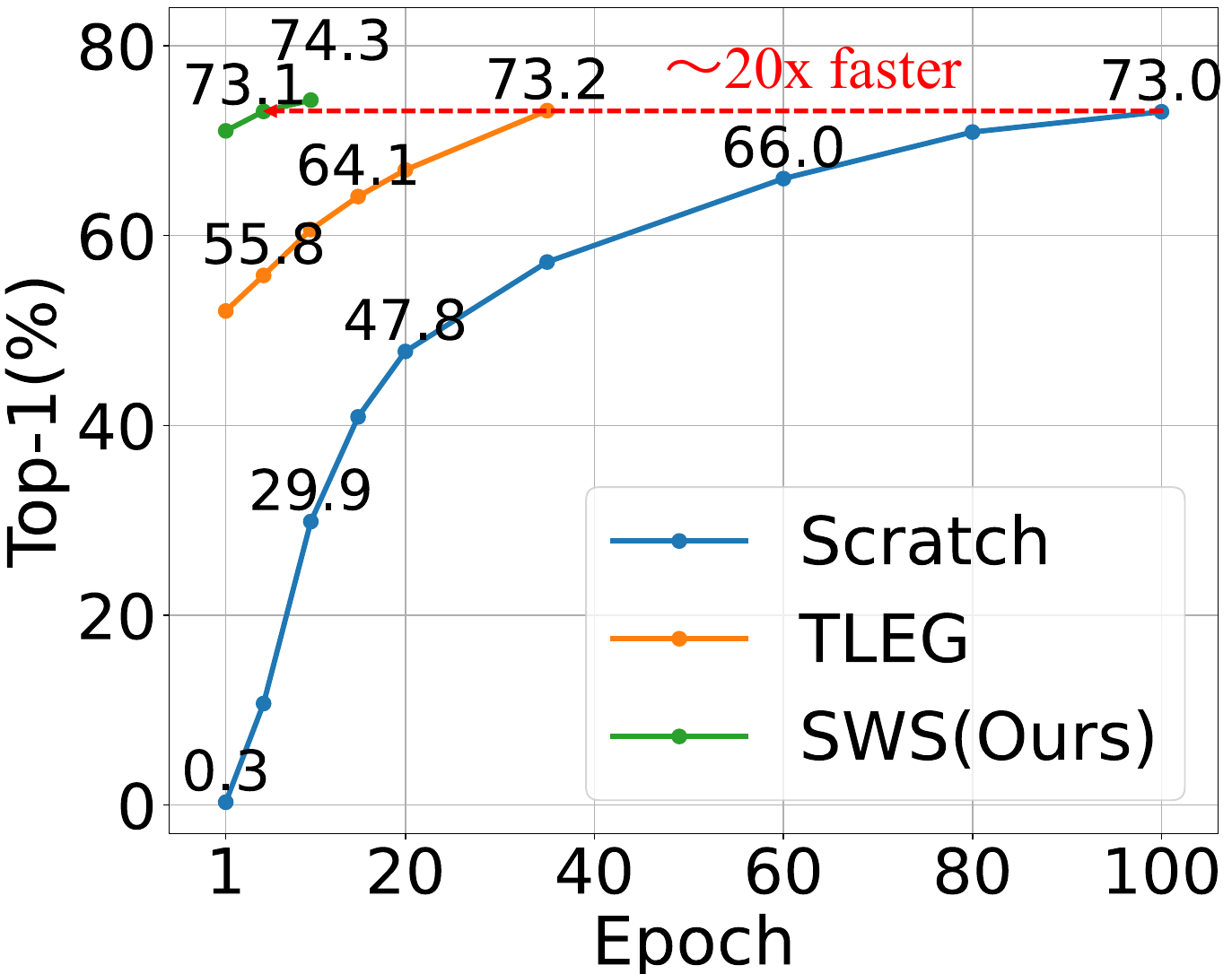}
        % \fbox{\rule{0pt}{1.1in} \rule{.9\linewidth}{0pt}}
        \caption{Des-S-9 (16.7M)}
        \label{fig5_4}
        \end{subfigure}
        \hfill
        \begin{subfigure}{0.195\linewidth}
        \includegraphics[scale=0.145]{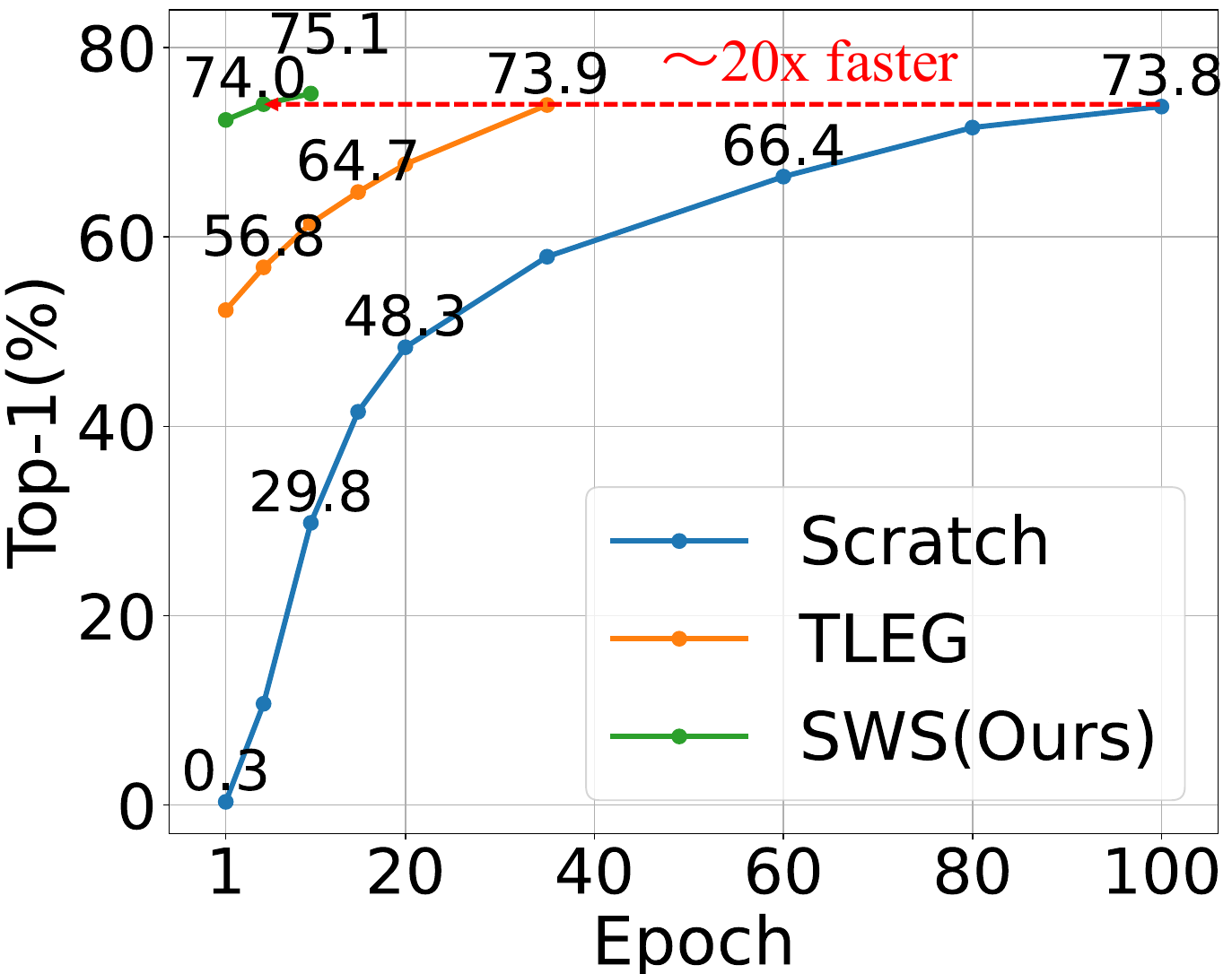}
        % \fbox{\rule{0pt}{1.1in} \rule{.9\linewidth}{0pt}}
        \caption{Des-S-10 (18.5M)}
        \label{fig5_5}
        \end{subfigure}
        \hfill

        \begin{subfigure}{0.195\linewidth}
        \includegraphics[scale=0.145]{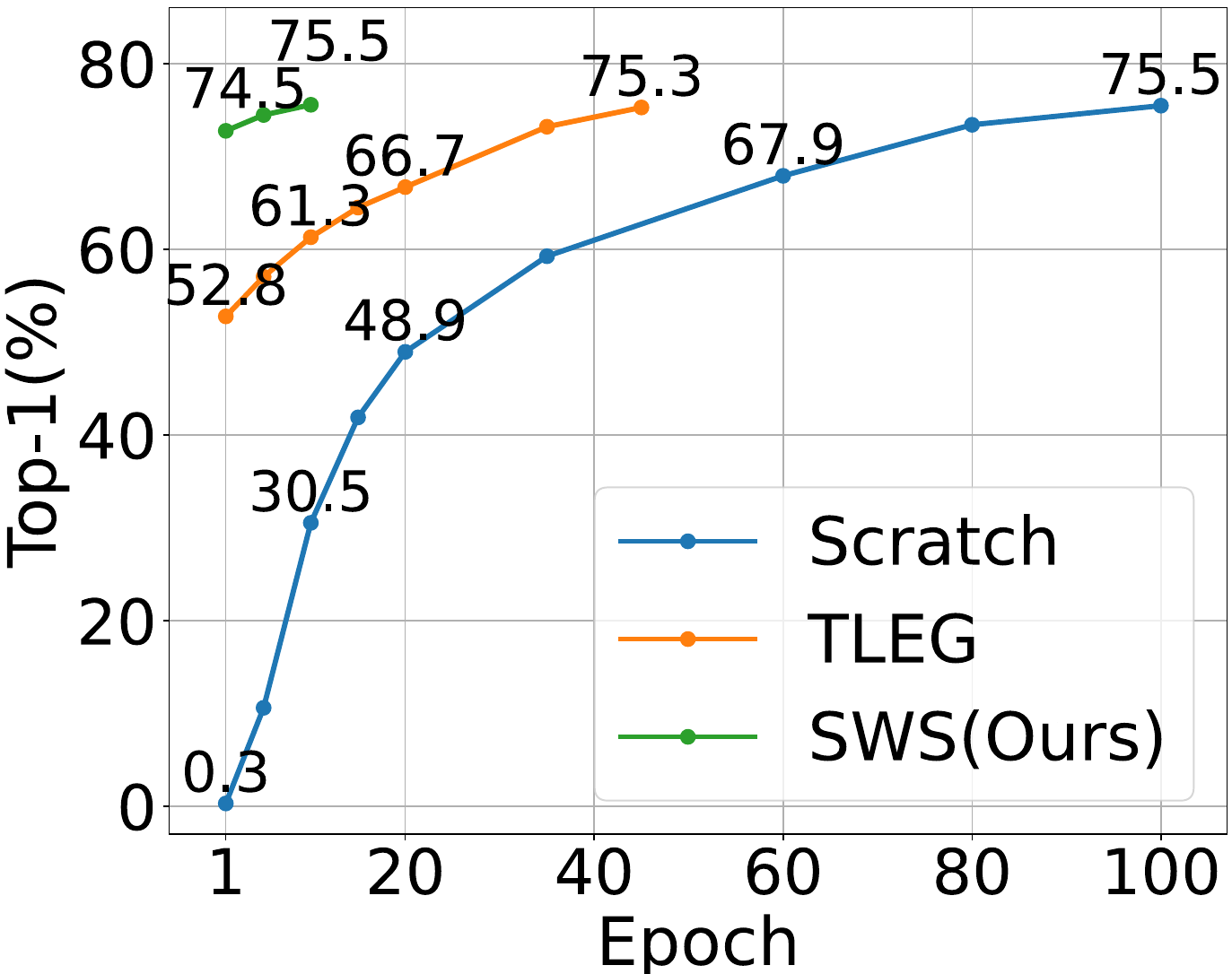}
        % \fbox{\rule{0pt}{1.1in} \rule{.9\linewidth}{0pt}}
        \caption{Des-S-11 (20.3M)}
        \label{fig5_6}
        \end{subfigure}
        \hfill
        \begin{subfigure}{0.195\linewidth}
        \includegraphics[scale=0.145]{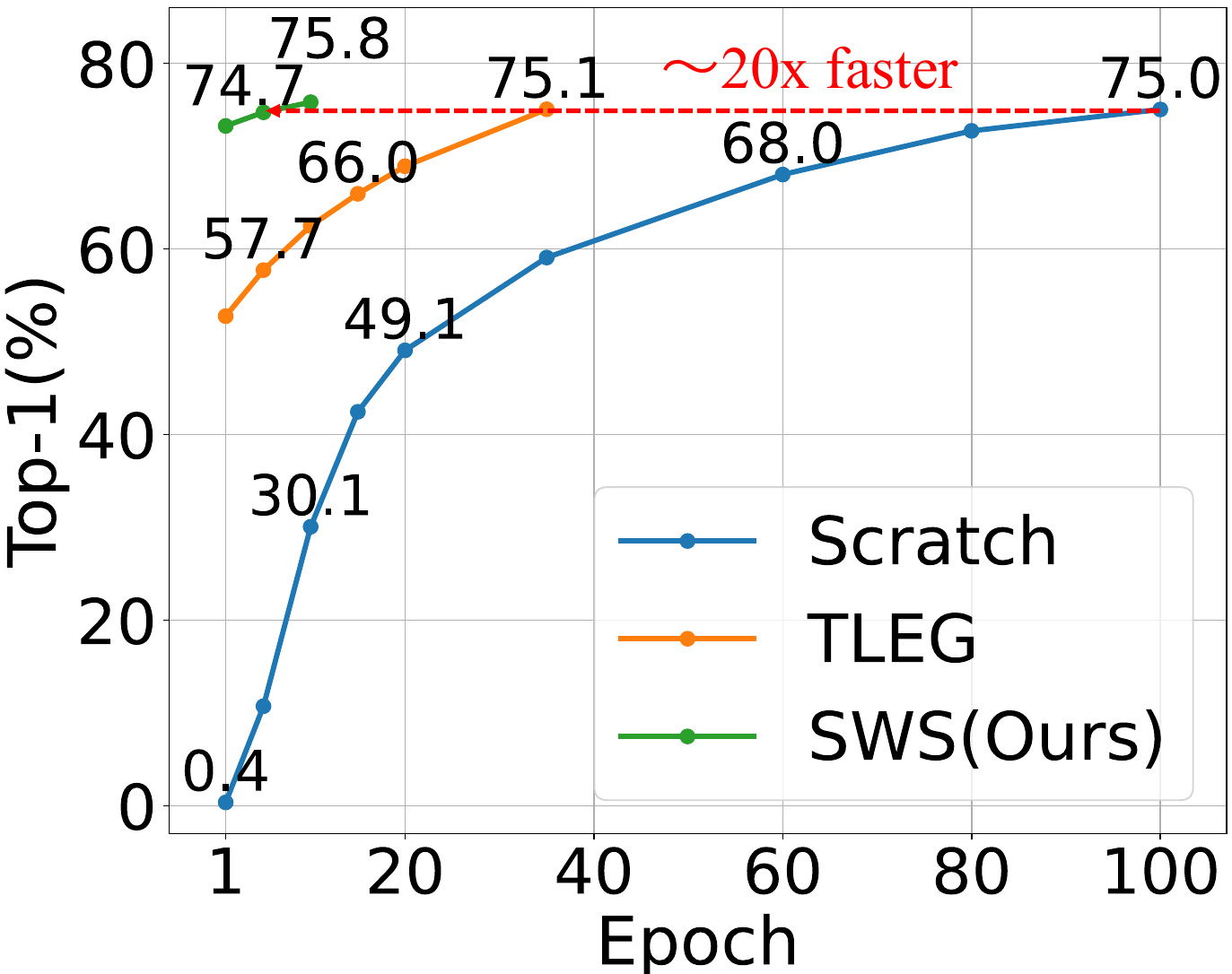}
        % \fbox{\rule{0pt}{1.1in} \rule{.9\linewidth}{0pt}}
        \caption{Des-S-12 (22.1M)}
        \label{fig5_7}
        \end{subfigure}
        \hfill
        \begin{subfigure}{0.195\linewidth}
        \includegraphics[scale=0.145]{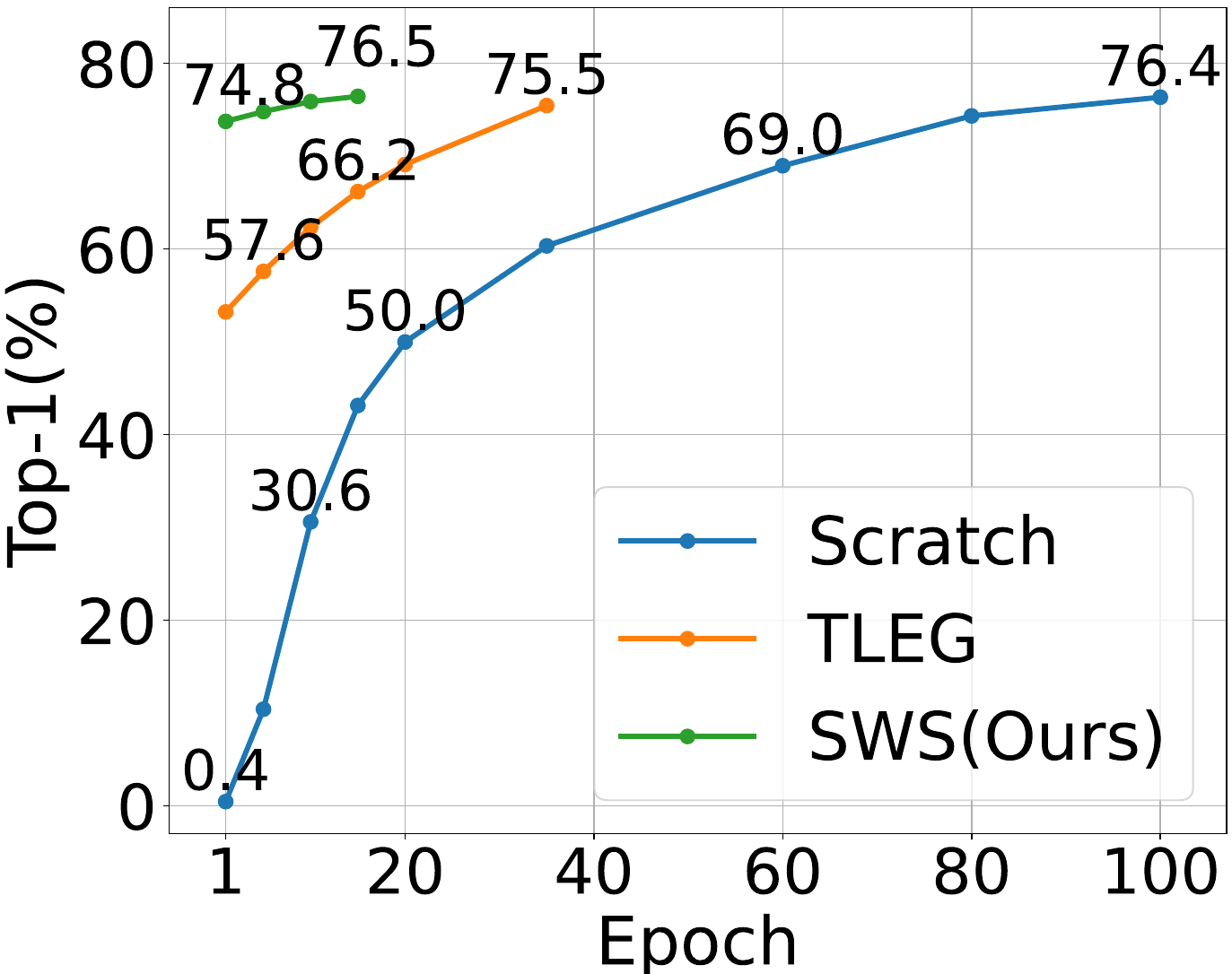}
        % \fbox{\rule{0pt}{1.1in} \rule{.9\linewidth}{0pt}}
        \caption{Des-S-13 (23.8M)}
        \label{fig5_8}
        \end{subfigure}
        \hfill
        \begin{subfigure}{0.195\linewidth}
        \includegraphics[scale=0.135]{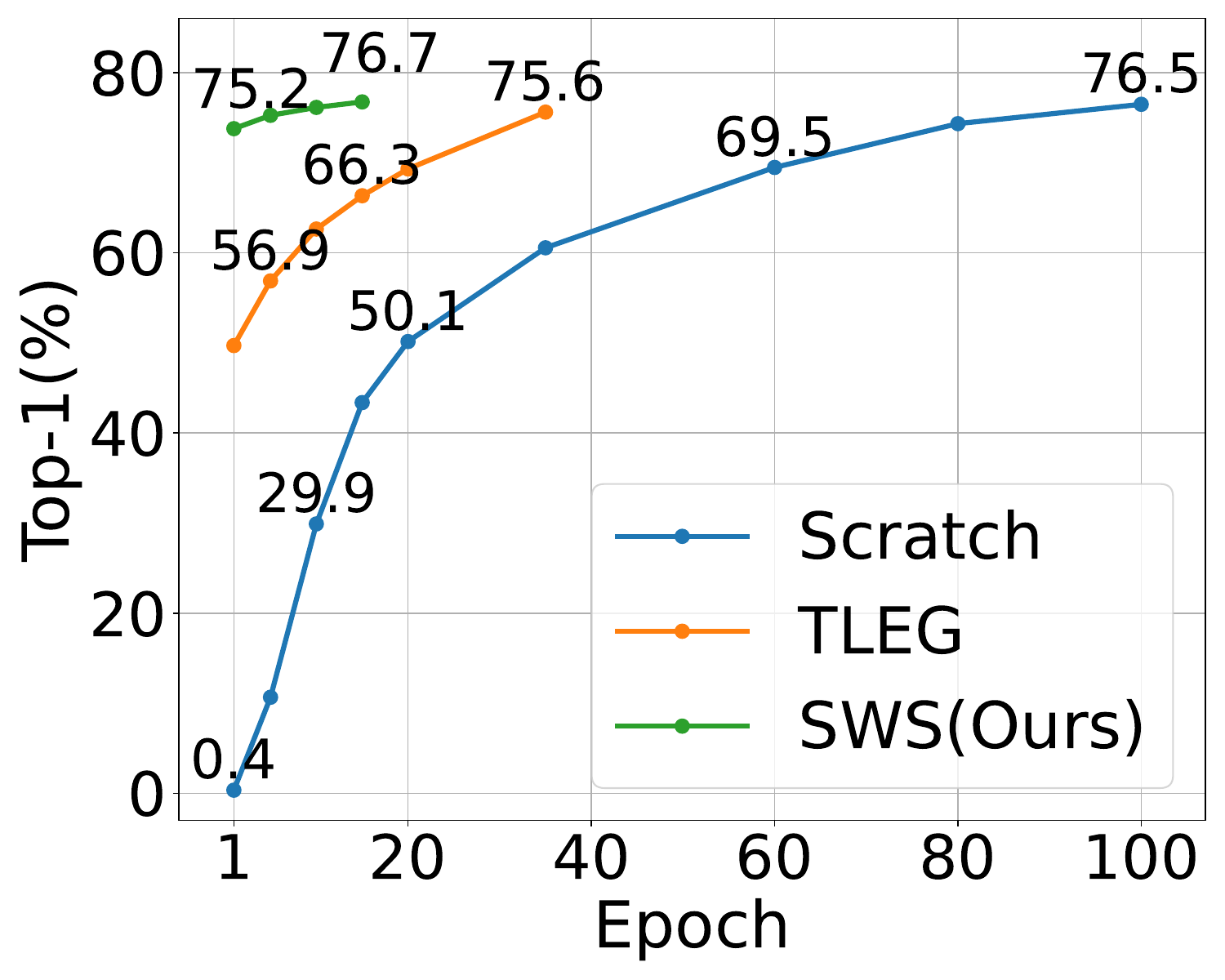}
        % \fbox{\rule{0pt}{1.1in} \rule{.9\linewidth}{0pt}}
        \caption{Des-S-14 (25.6M)}
        \label{fig5_9}
        \end{subfigure}
        \hfill
        \begin{subfigure}{0.195\linewidth}
        \includegraphics[scale=0.145]{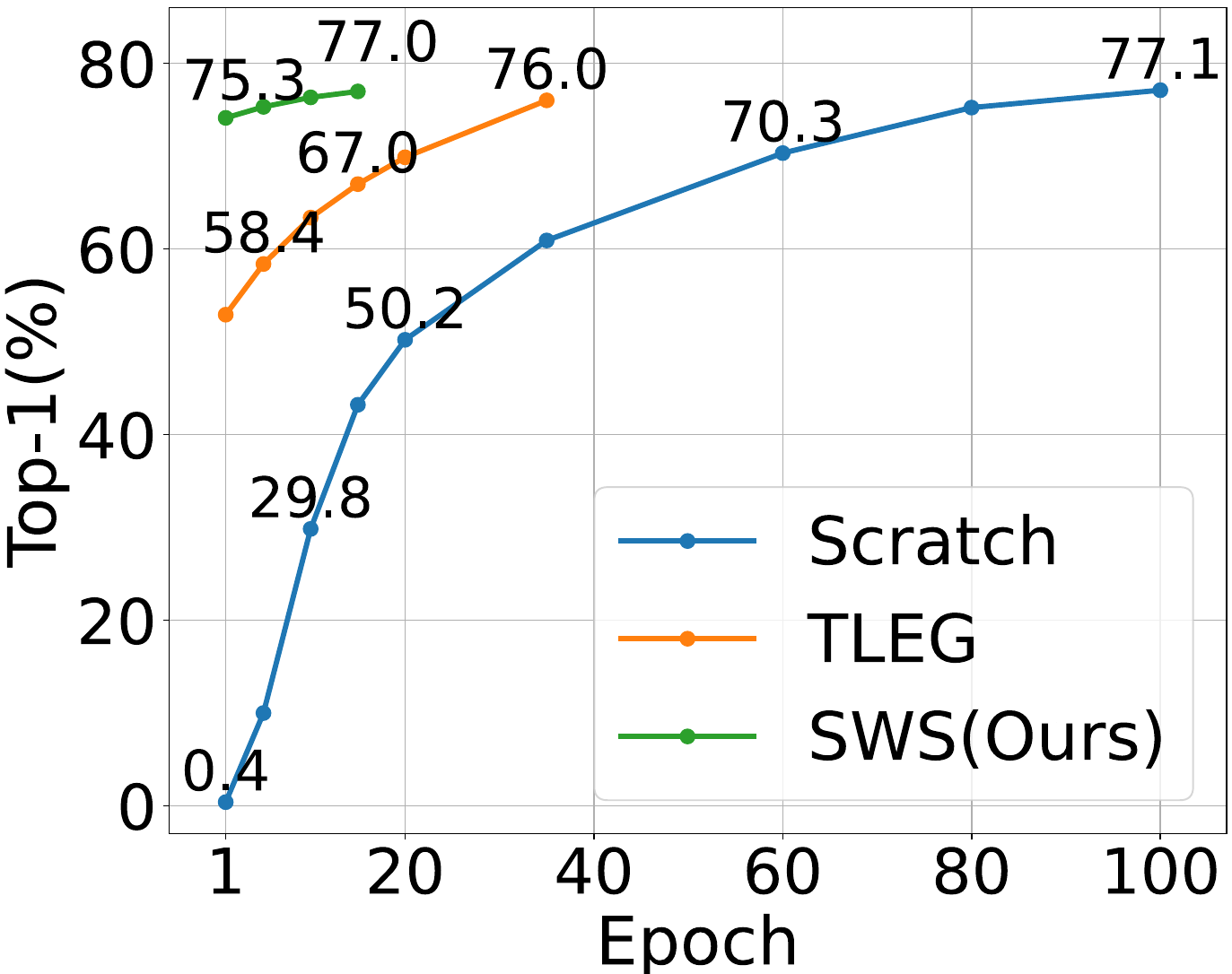}
        % \fbox{\rule{0pt}{1.1in} \rule{.9\linewidth}{0pt}}
        \caption{Des-S-15 (27.4M)}
        \label{fig5_10}
        \end{subfigure}
        \hfill

        \begin{subfigure}{0.195\linewidth}
        \includegraphics[scale=0.145]{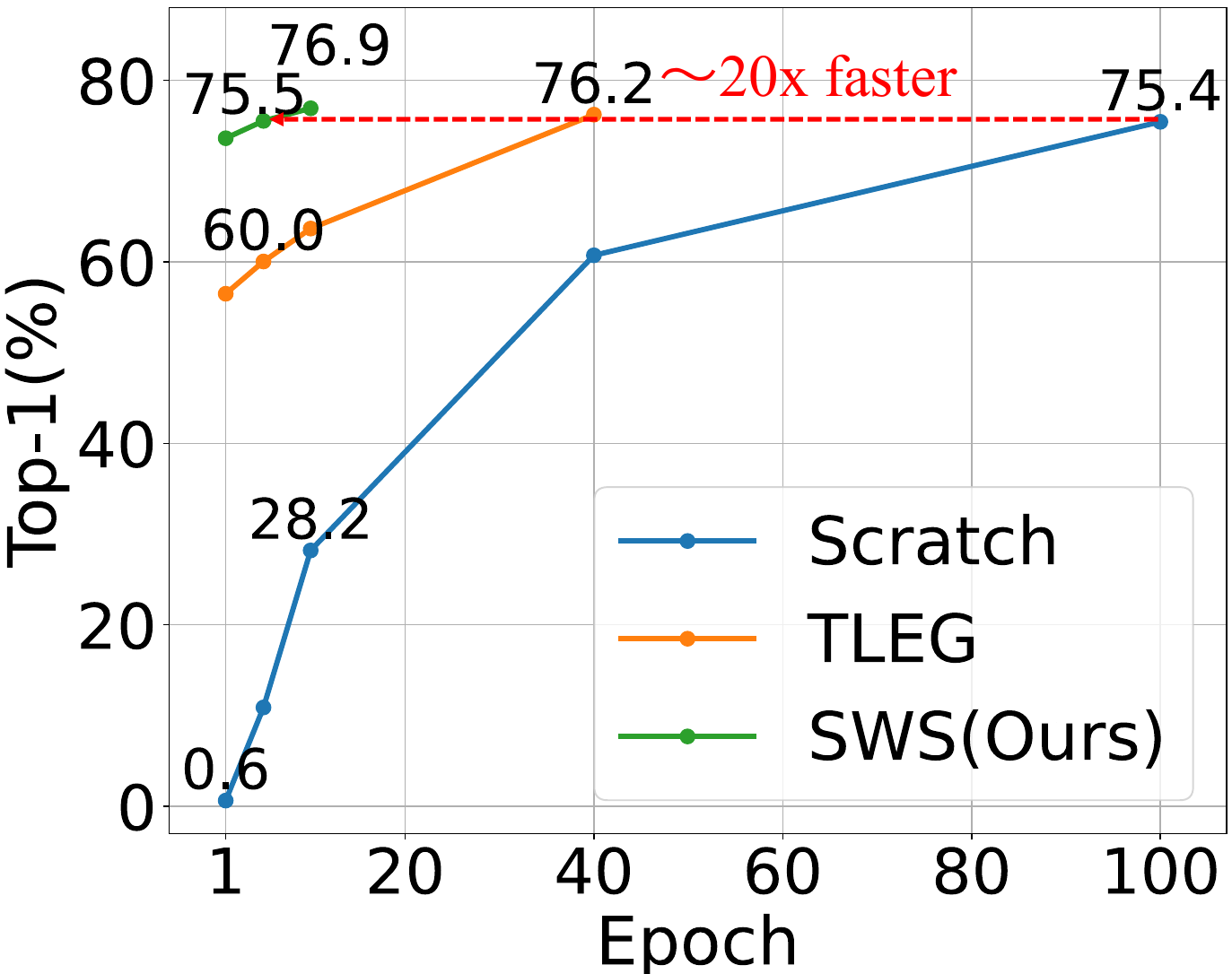}
        % \fbox{\rule{0pt}{1.1in} \rule{.9\linewidth}{0pt}}
        \caption{Des-B-6 (44.0M)}
        \label{fig5_11}
        \end{subfigure}
        \hfill
        \begin{subfigure}{0.195\linewidth}
        \includegraphics[scale=0.145]{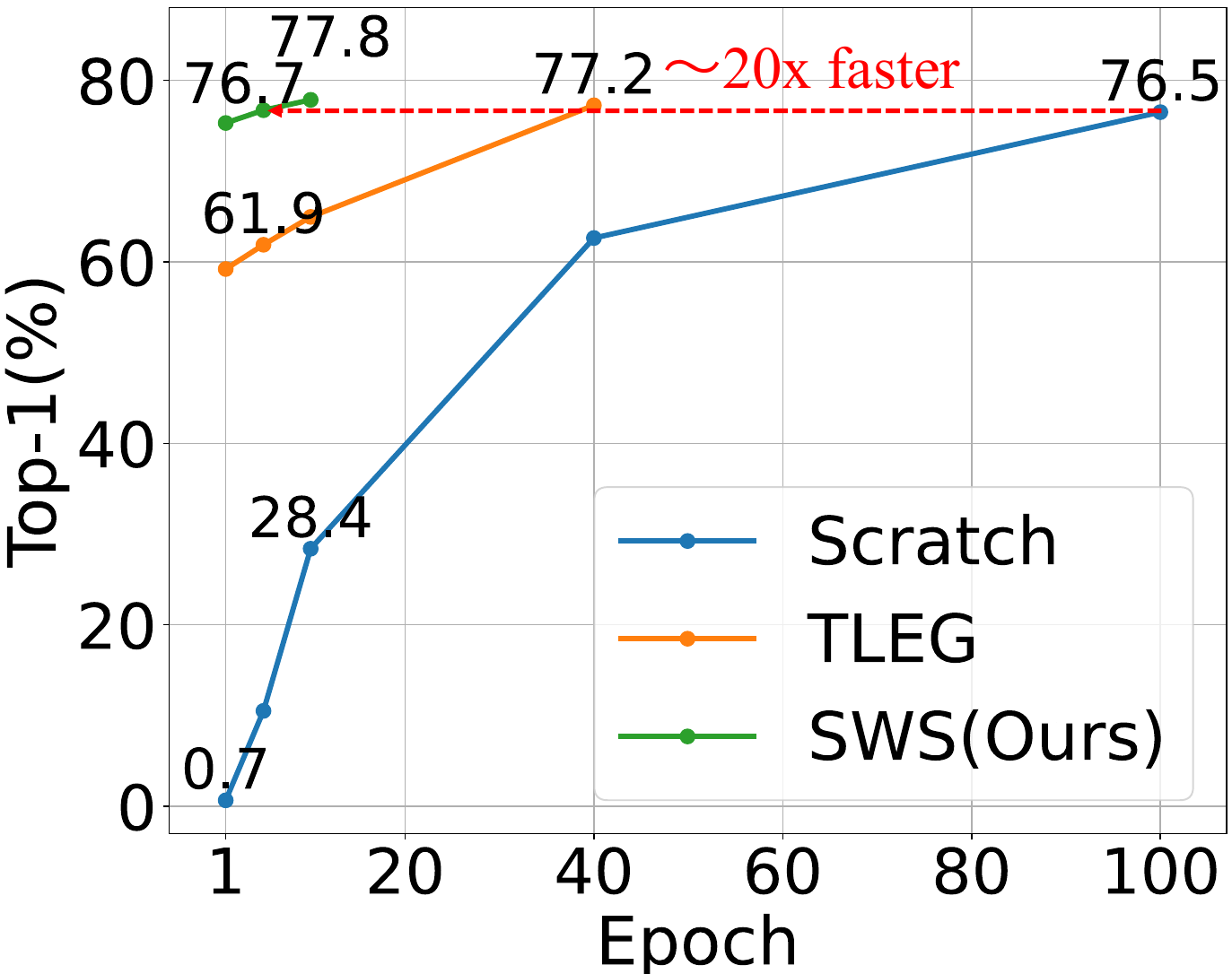}
        % \fbox{\rule{0pt}{1.1in} \rule{.9\linewidth}{0pt}}
        \caption{Des-B-7 (51.1M)}
        \label{fig5_12}
        \end{subfigure}
        \hfill
        \begin{subfigure}{0.195\linewidth}
        \includegraphics[scale=0.145]{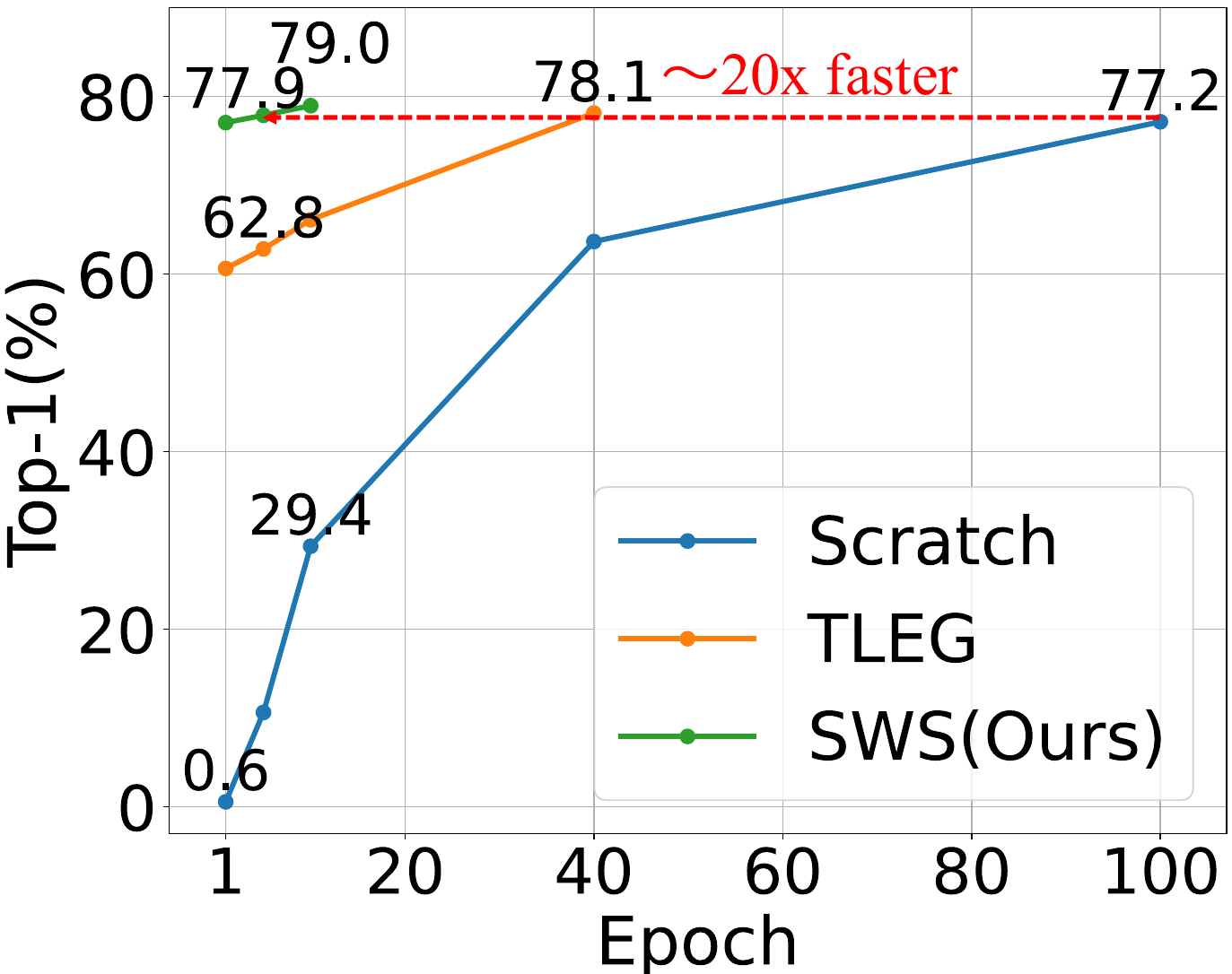}
        % \fbox{\rule{0pt}{1.1in} \rule{.9\linewidth}{0pt}}
        \caption{Des-B-8 (58.2M)}
        \label{fig5_13}
        \end{subfigure}
        \hfill
        \begin{subfigure}{0.195\linewidth}
        \includegraphics[scale=0.145]{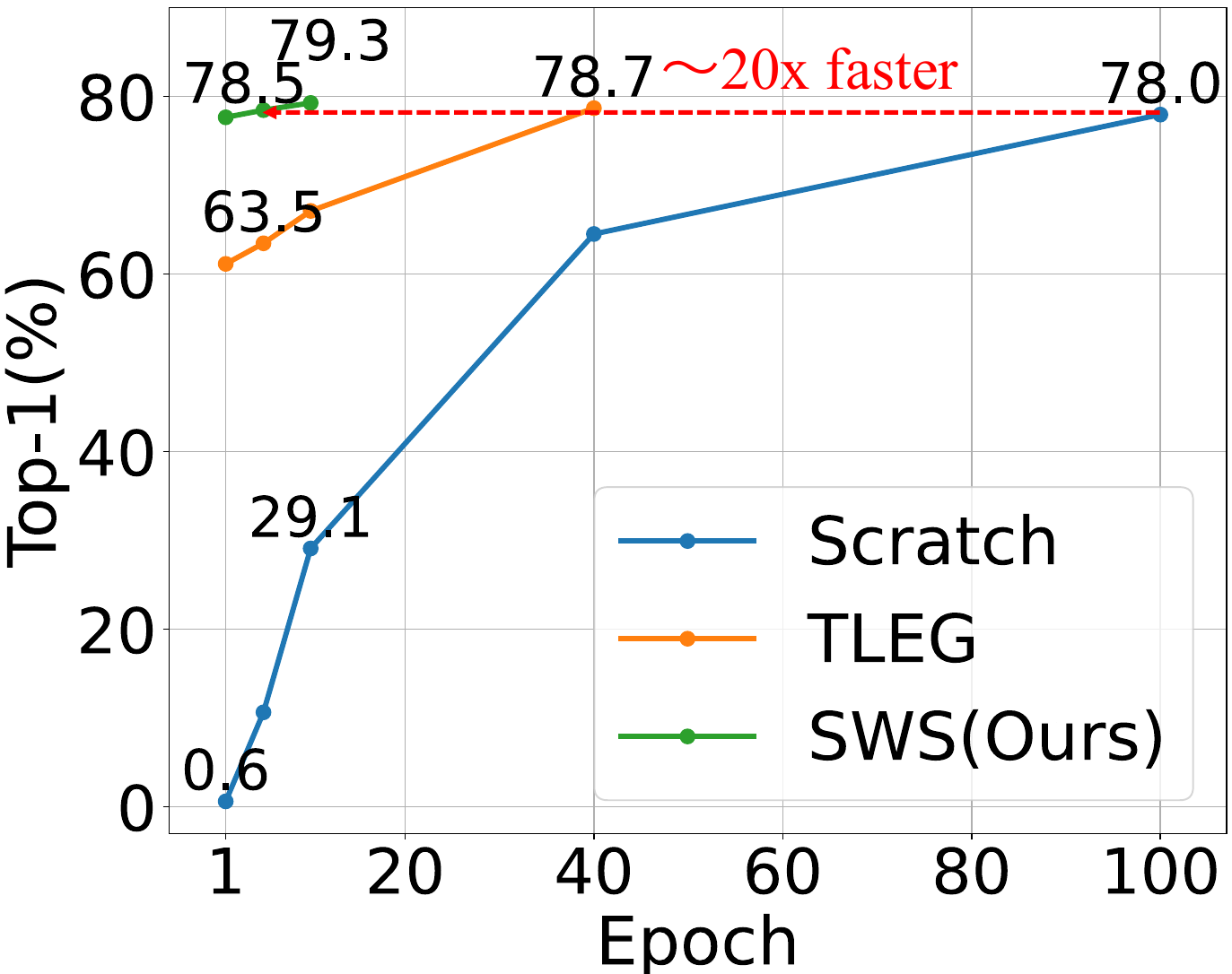}
        % \fbox{\rule{0pt}{1.1in} \rule{.9\linewidth}{0pt}}
        \caption{Des-B-9 (65.3M)}
        \label{fig5_14}
        \end{subfigure}
        \hfill
        \begin{subfigure}{0.195\linewidth}
        \includegraphics[scale=0.145]{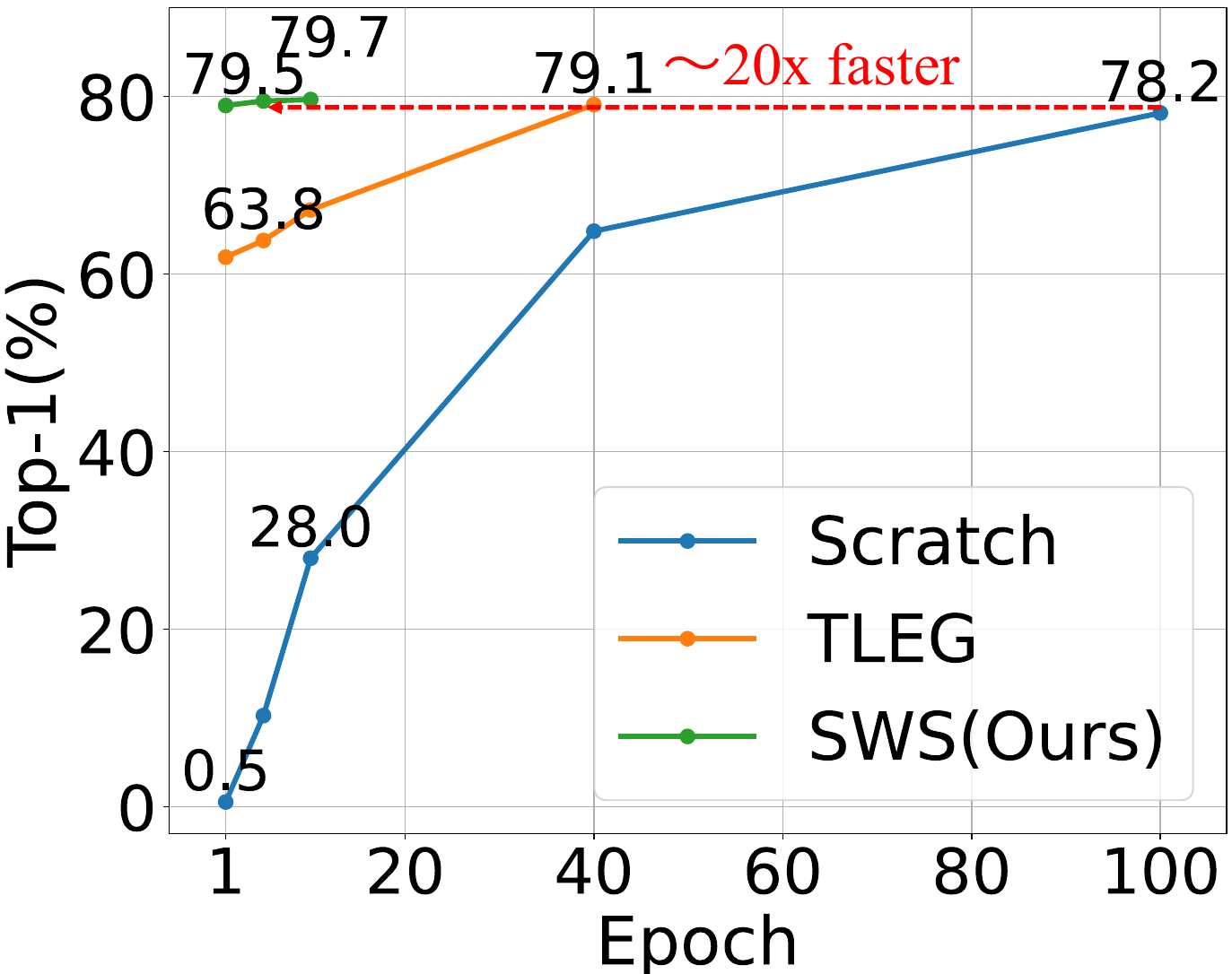}
        % \fbox{\rule{0pt}{1.1in} \rule{.9\linewidth}{0pt}}
        \caption{Des-B-10 (72.4M)}
        \label{fig5_15}
        \end{subfigure}
        \hfill

        \begin{subfigure}{0.195\linewidth}
        \includegraphics[scale=0.145]{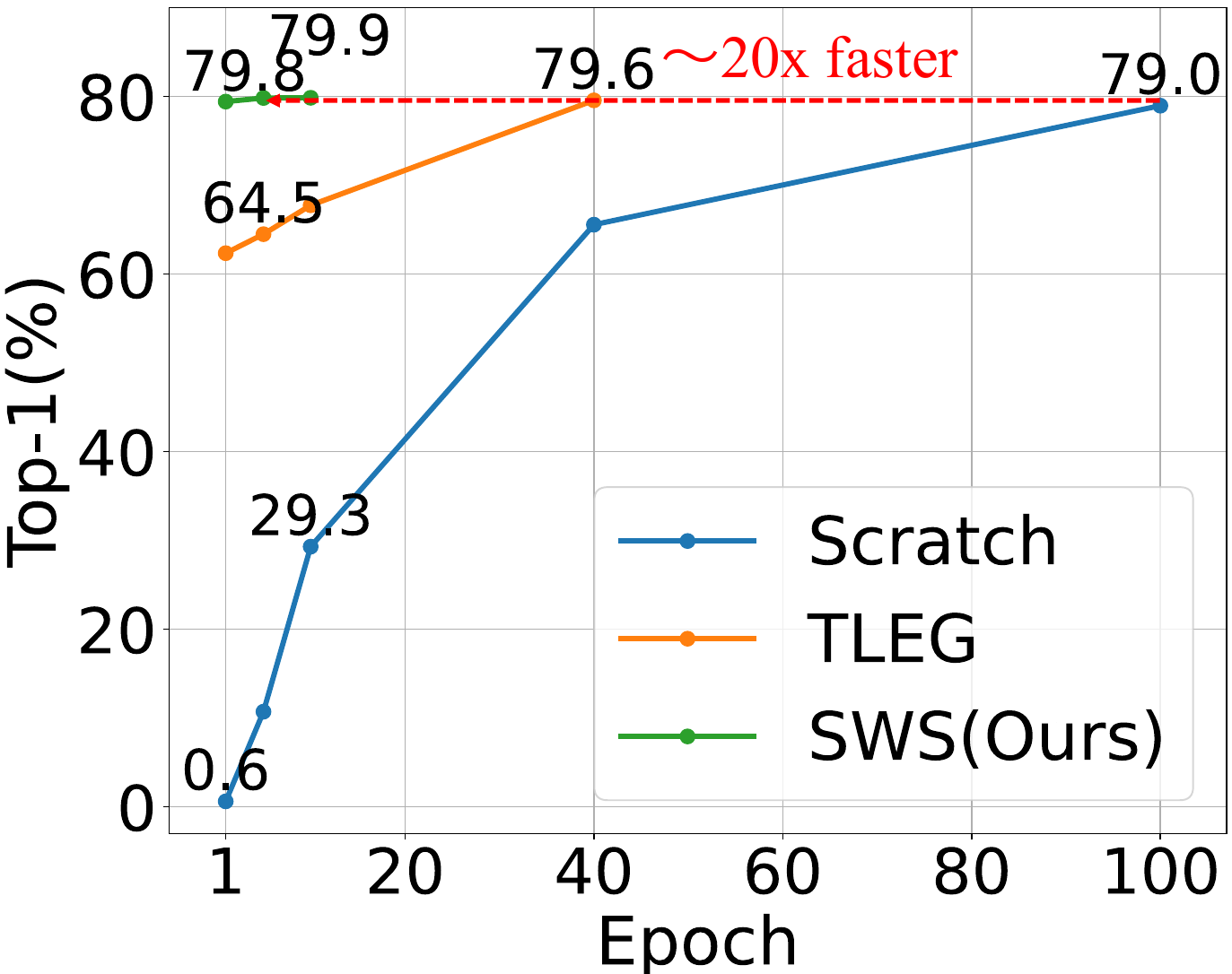}
        % \fbox{\rule{0pt}{1.1in} \rule{.9\linewidth}{0pt}}
        \caption{Des-B-11 (79.5M)}
        \label{fig5_16}
        \end{subfigure}
        \hfill
        \begin{subfigure}{0.195\linewidth}
        \includegraphics[scale=0.145]{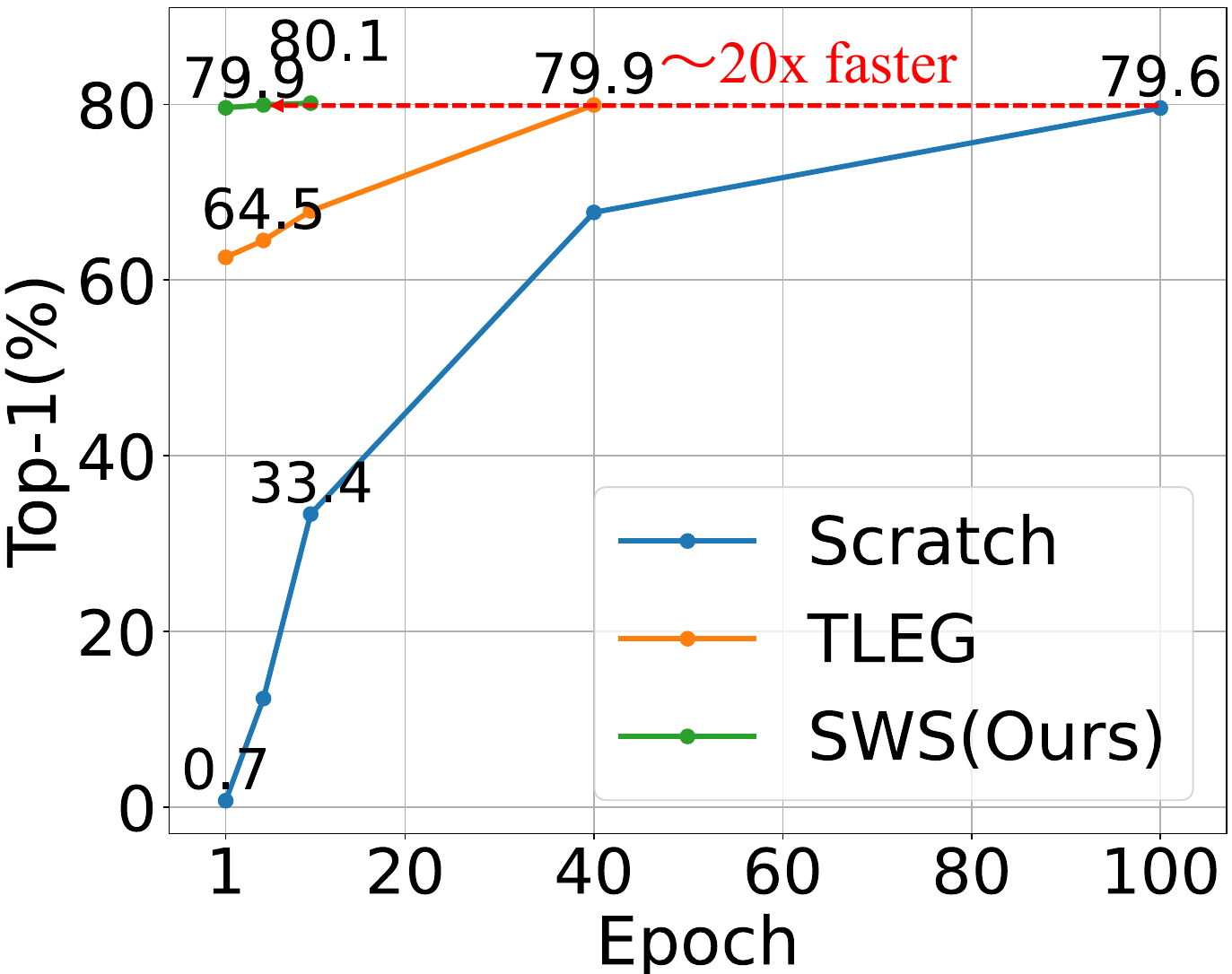}
        % \fbox{\rule{0pt}{1.1in} \rule{.9\linewidth}{0pt}}
        \caption{Des-B-12 (86.6M)}
        \label{fig5_17}
        \end{subfigure}
        \hfill
        \begin{subfigure}{0.195\linewidth}
        \includegraphics[scale=0.145]{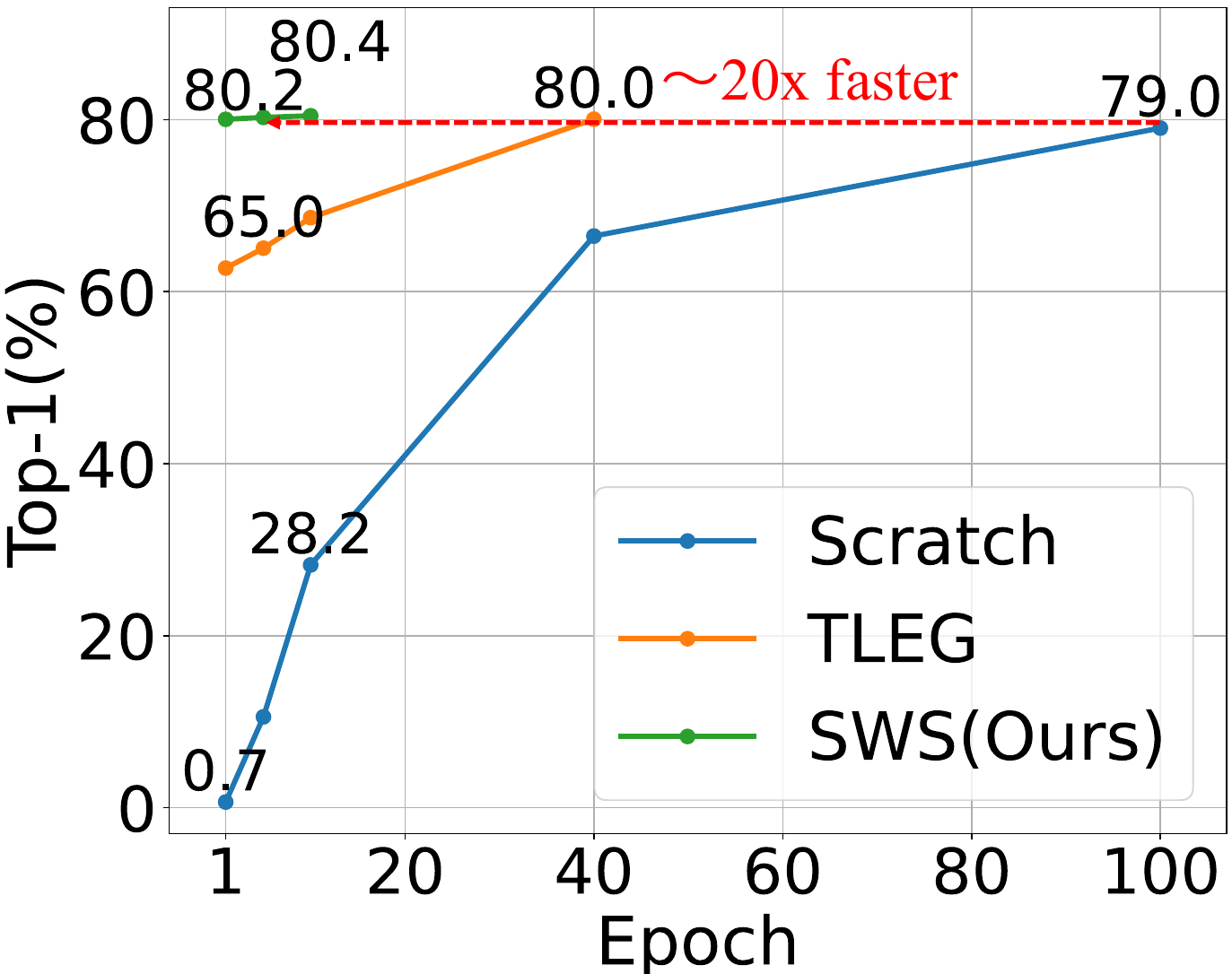}
        % \fbox{\rule{0pt}{1.1in} \rule{.9\linewidth}{0pt}}
        \caption{Des-B-13 (93.7M)}
        \label{fig5_18}
        \end{subfigure}
        \hfill
        \begin{subfigure}{0.195\linewidth}
        \includegraphics[scale=0.145]{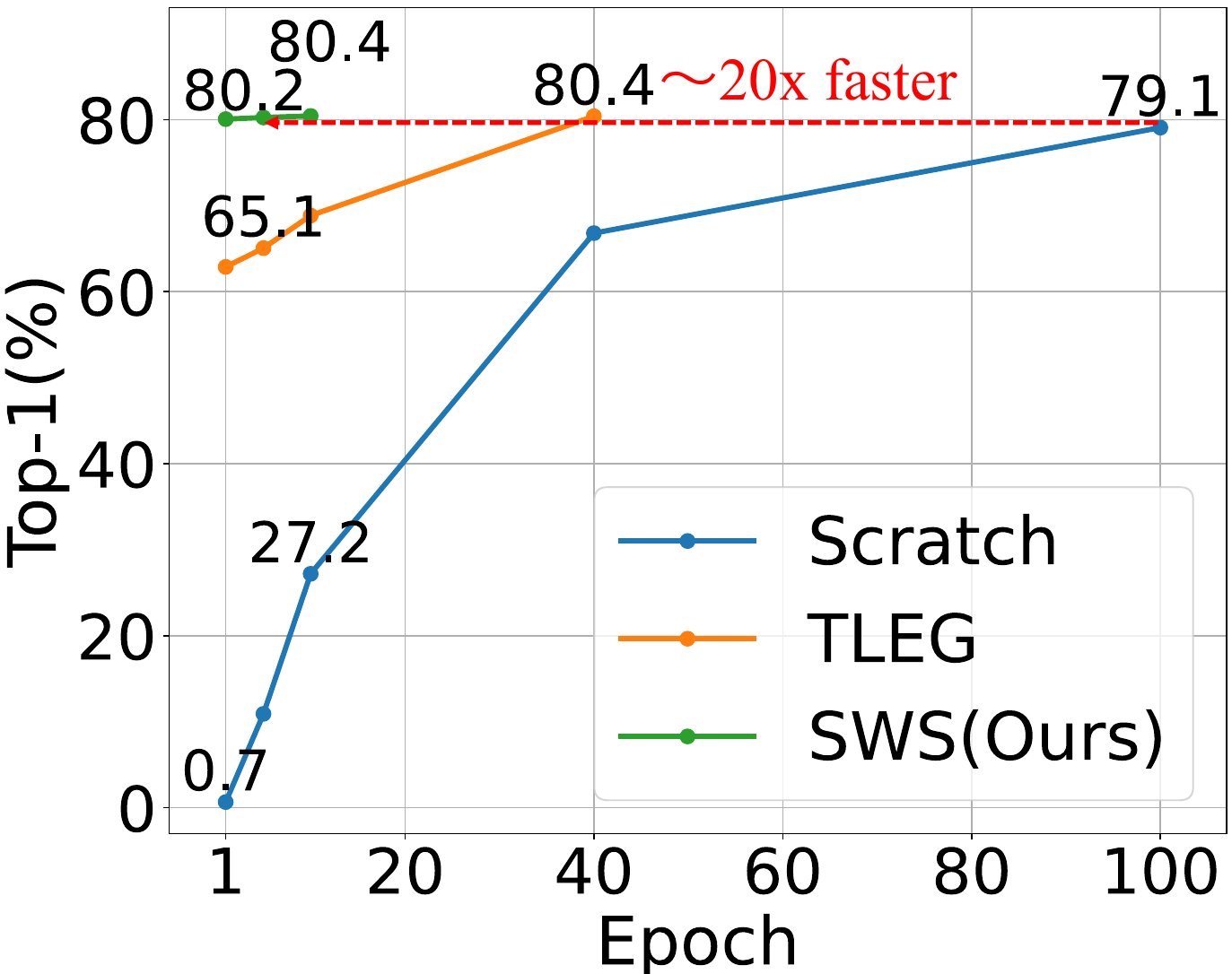}
        % \fbox{\rule{0pt}{1.1in} \rule{.9\linewidth}{0pt}}
        \caption{Des-B-14 (100.7M)}
        \label{fig5_19}
        \end{subfigure}
        \hfill
        \begin{subfigure}{0.195\linewidth}
        \includegraphics[scale=0.145]{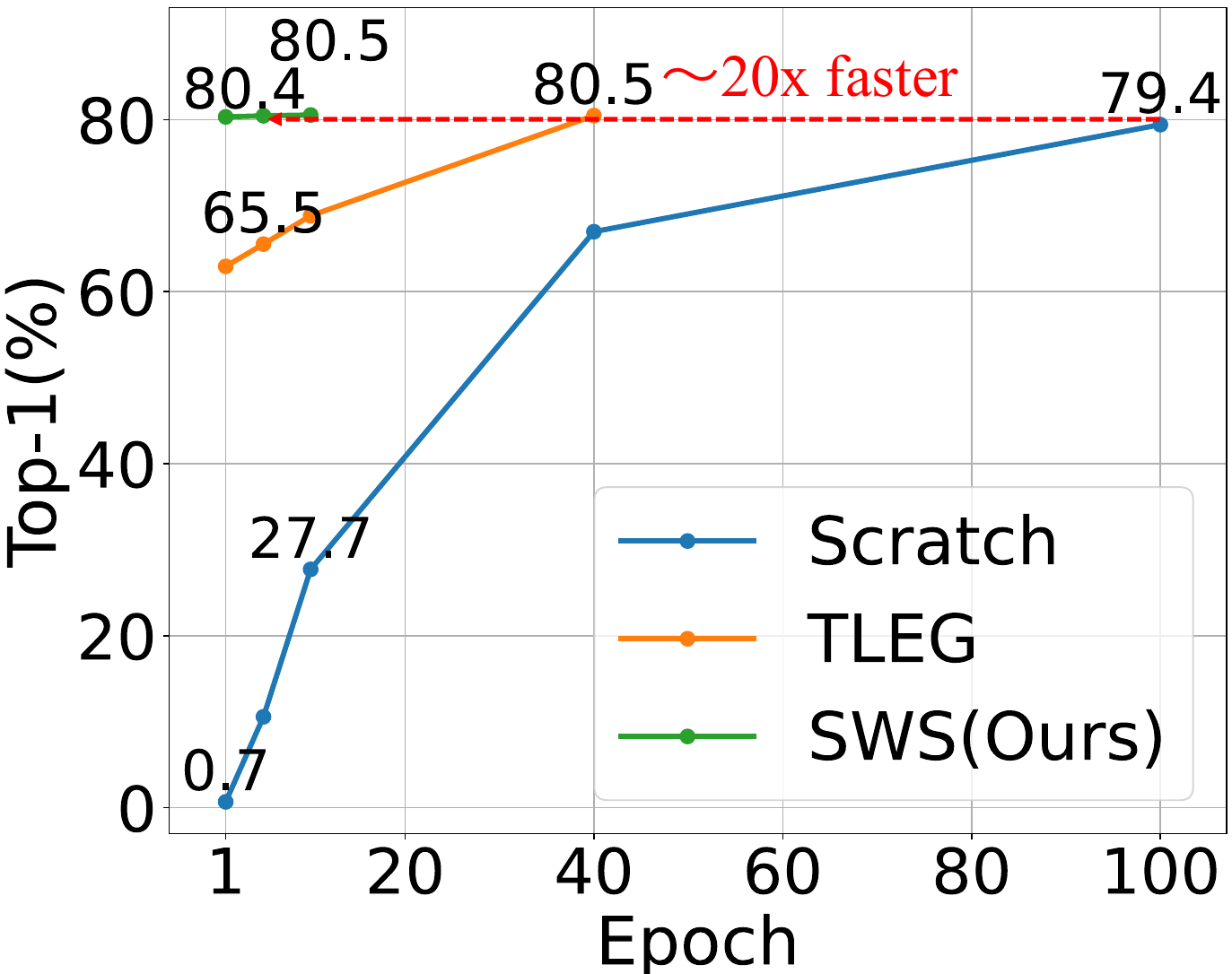}
        % \fbox{\rule{0pt}{1.1in} \rule{.9\linewidth}{0pt}}
        \caption{Des-B-15 (107.8M)}
        \label{fig5_20}
        \end{subfigure}
        \hfill
        
    \caption{ Performance comparisons on ImageNet-1K between several baselines and SWS.
    Number in bracket of (a)-(t) means Params(M).
    }
    % \vspace{-3pt}
    \label{fig:im1k_result1}
\end{figure*}

\begin{figure*}[ht]
    \centering
        \begin{subfigure}{0.245\linewidth}
        \includegraphics[scale=0.17]{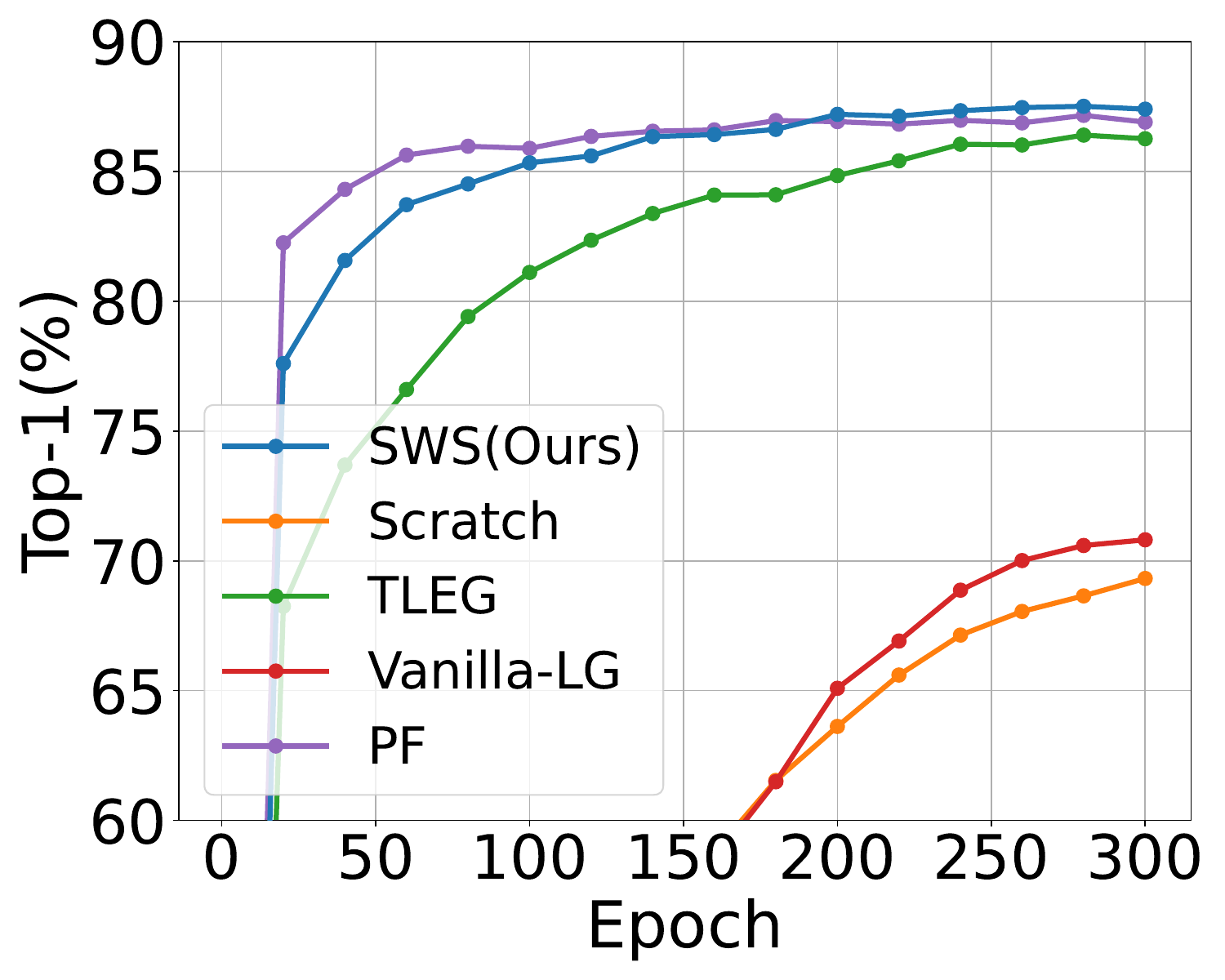}
        % \fbox{\rule{0pt}{1.1in} \rule{.9\linewidth}{0pt}}
        \caption{CIFAR-100}
        \label{fig6_1}
        \end{subfigure}
        \hfill
        \begin{subfigure}{0.245\linewidth}
        \includegraphics[scale=0.17]{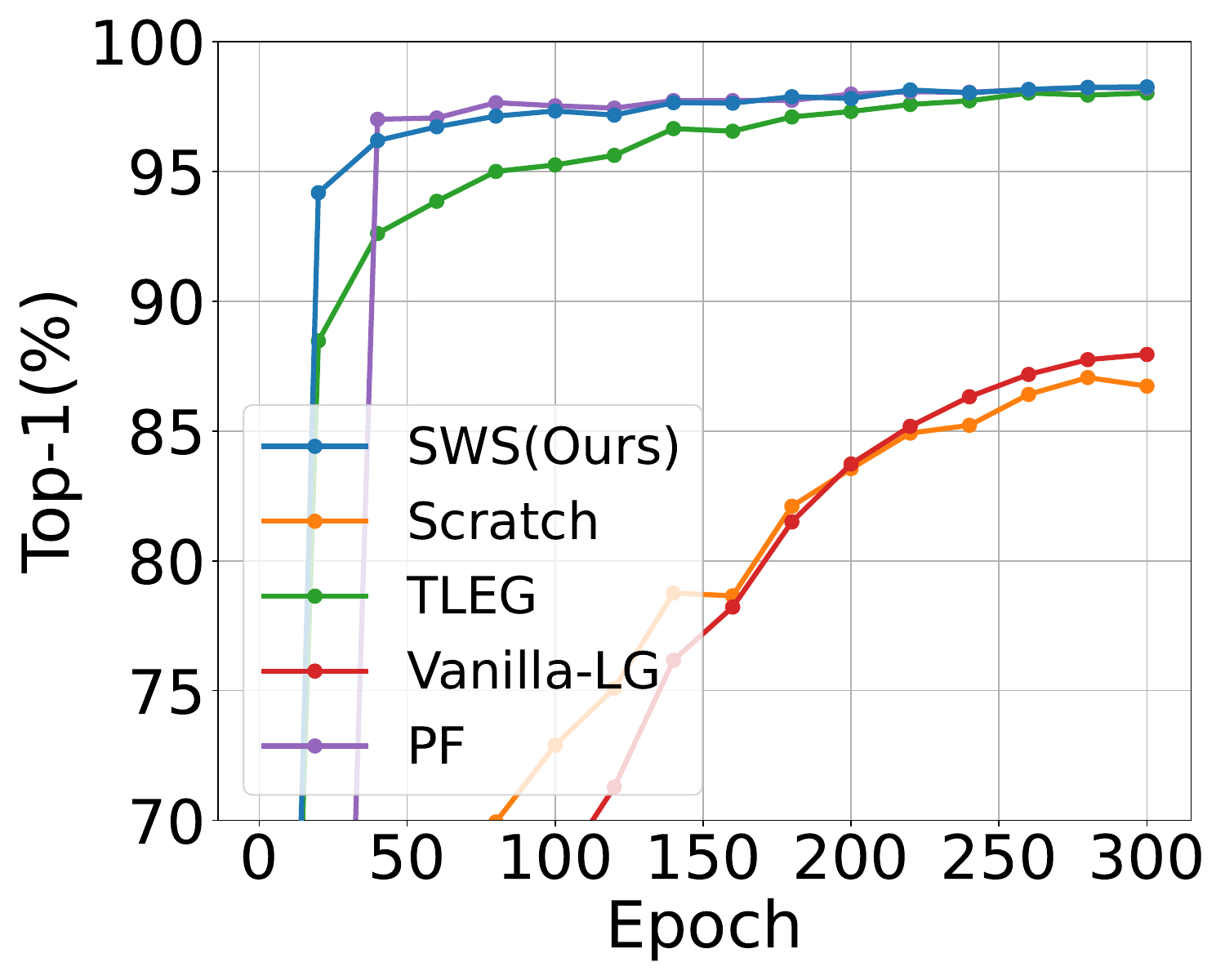}
        % \fbox{\rule{0pt}{1.1in} \rule{.9\linewidth}{0pt}}
        \caption{CIFAR-10}
        \label{fig6_2}
        \end{subfigure}
        \hfill
        \begin{subfigure}{0.245\linewidth}
        \includegraphics[scale=0.17]{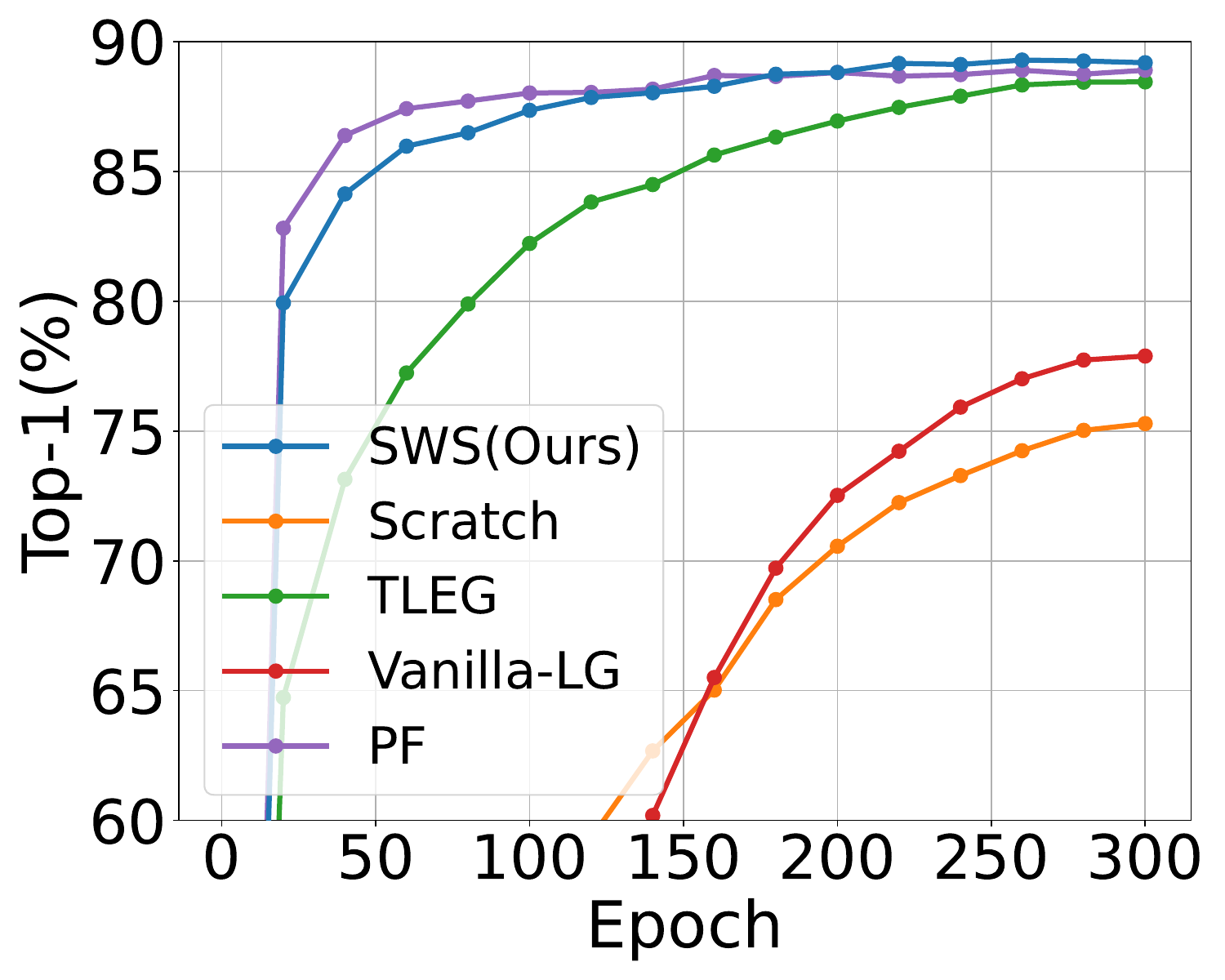}
        % \fbox{\rule{0pt}{1.1in} \rule{.9\linewidth}{0pt}}
        \caption{Food-101}
        \label{fig6_3}
        \end{subfigure}
        \hfill
        \begin{subfigure}{0.245\linewidth}
        \includegraphics[scale=0.17]{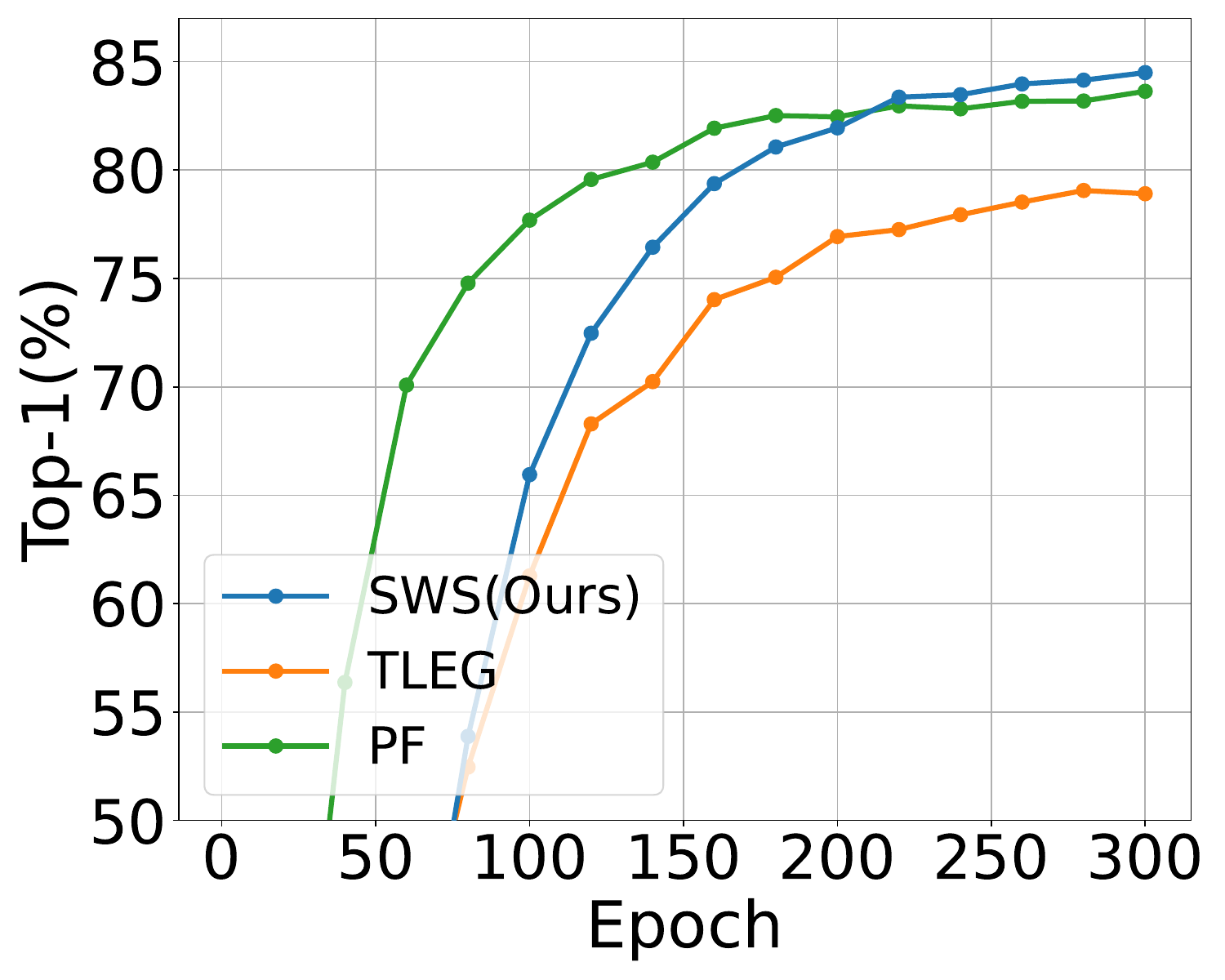}
        % \fbox{\rule{0pt}{1.1in} \rule{.9\linewidth}{0pt}}
        \caption{Cars-196}
        \label{fig6_4}
        \end{subfigure}

        \begin{subfigure}{0.245\linewidth}
        \includegraphics[scale=0.17]{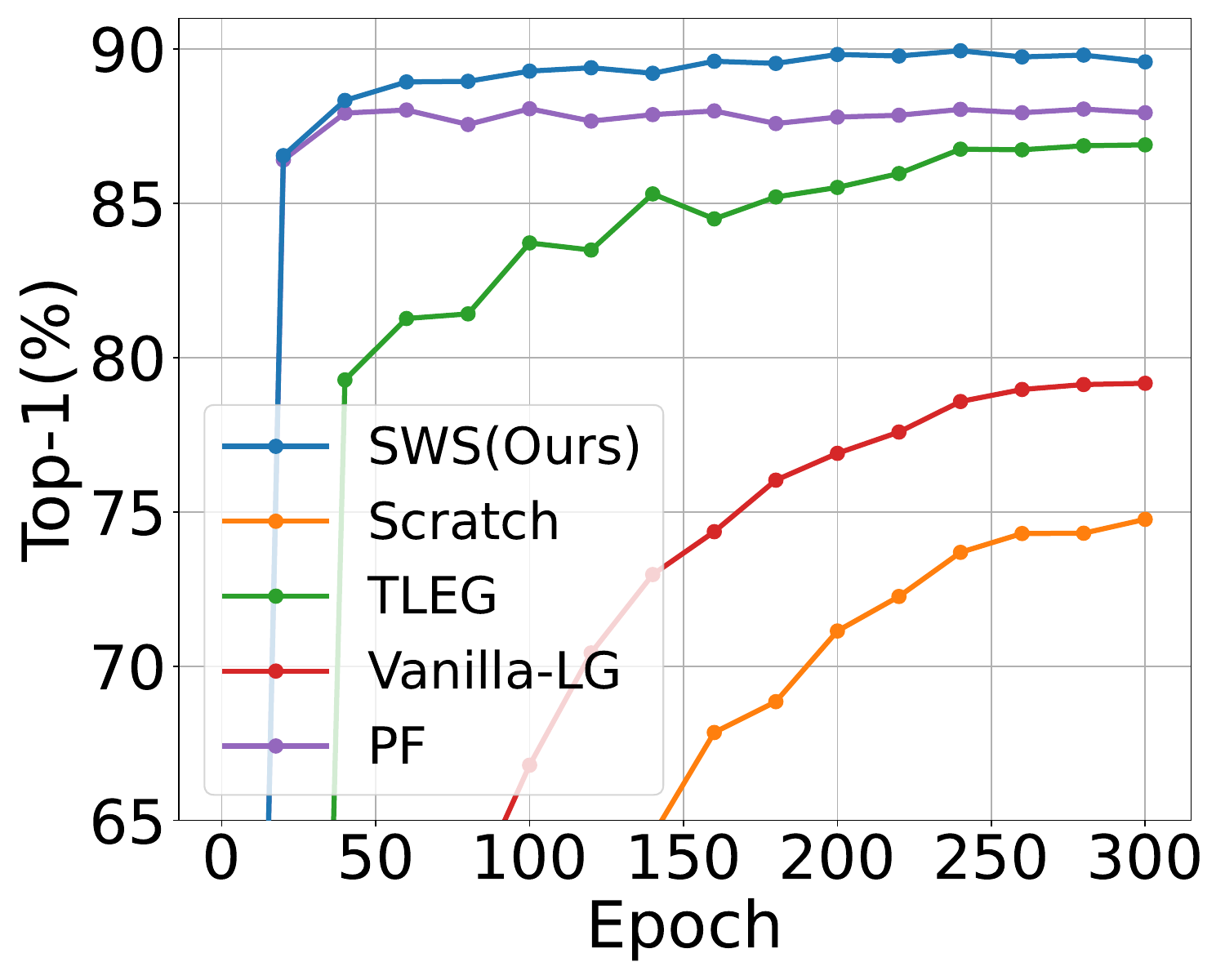}
        % \fbox{\rule{0pt}{1.1in} \rule{.9\linewidth}{0pt}}
        \caption{CIFAR-100}
        \label{fig6_5}
        \end{subfigure}
        \hfill
        \begin{subfigure}{0.245\linewidth}
        \includegraphics[scale=0.17]{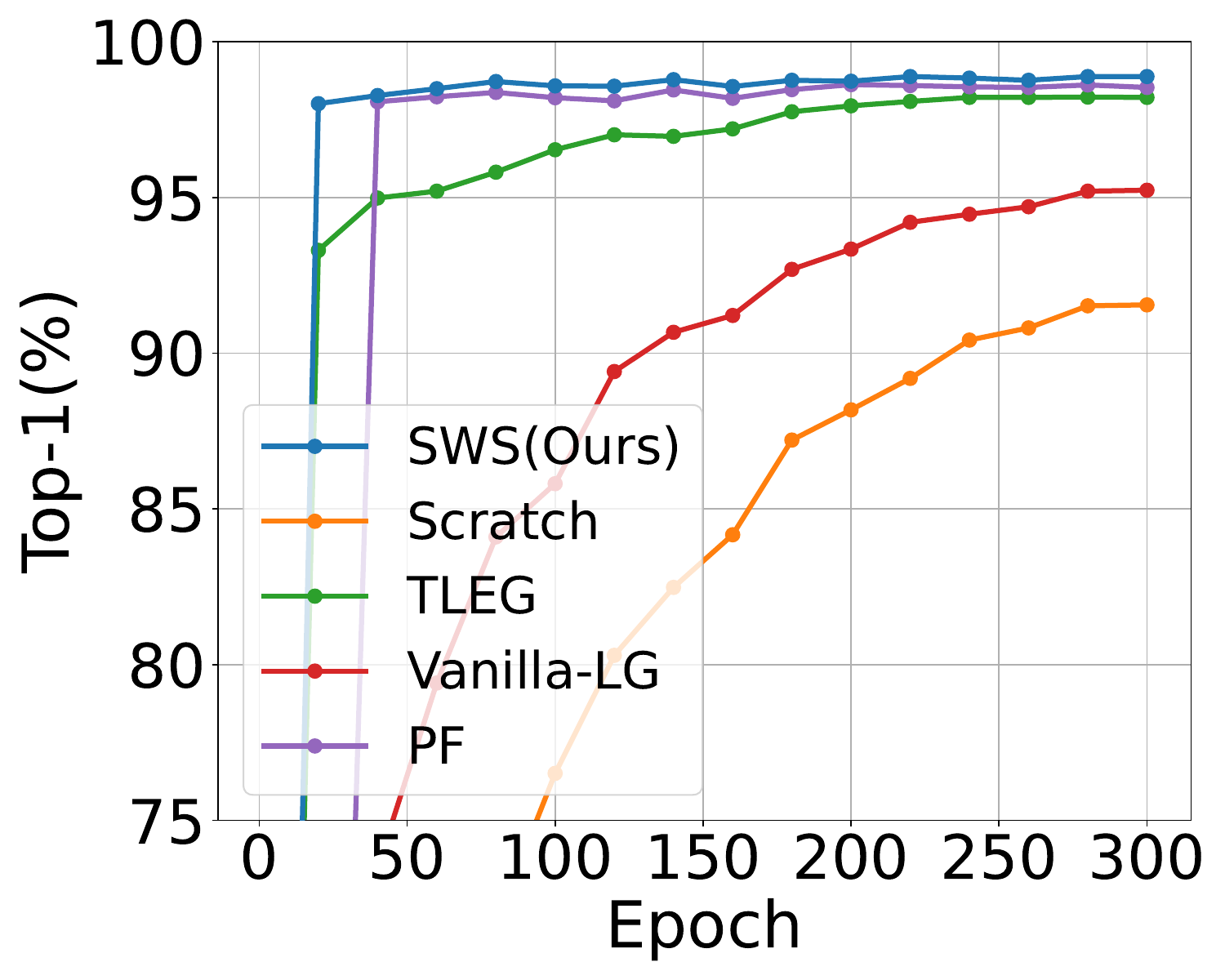}
        % \fbox{\rule{0pt}{1.1in} \rule{.9\linewidth}{0pt}}
        \caption{CIFAR-10}
        \label{fig6_6}
        \end{subfigure}
        \hfill
        \begin{subfigure}{0.245\linewidth}
        \includegraphics[scale=0.17]{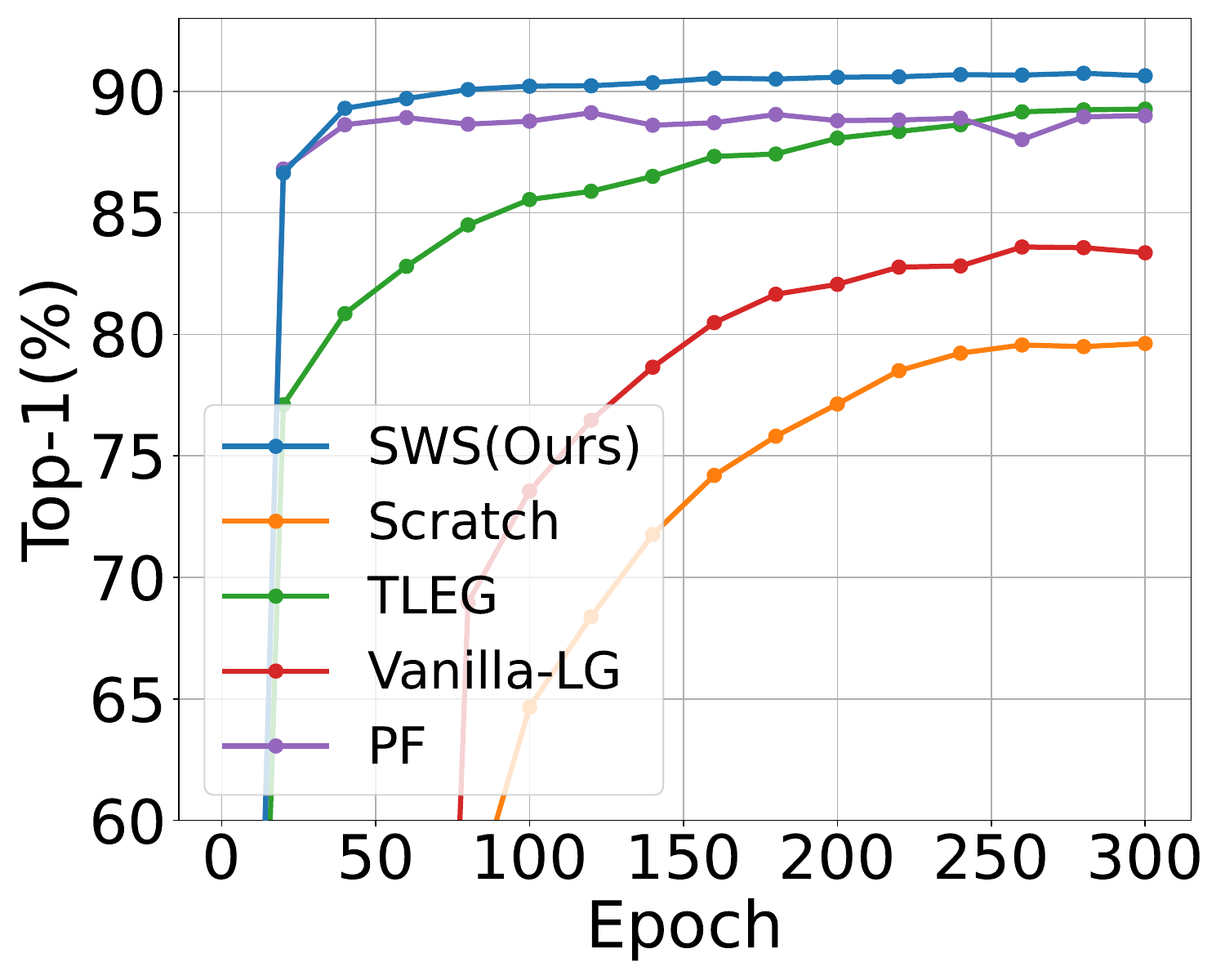}
        % \fbox{\rule{0pt}{1.1in} \rule{.9\linewidth}{0pt}}
        \caption{Food-101}
        \label{fig6_7}
        \end{subfigure}
        \hfill
        \begin{subfigure}{0.245\linewidth}
        \includegraphics[scale=0.17]{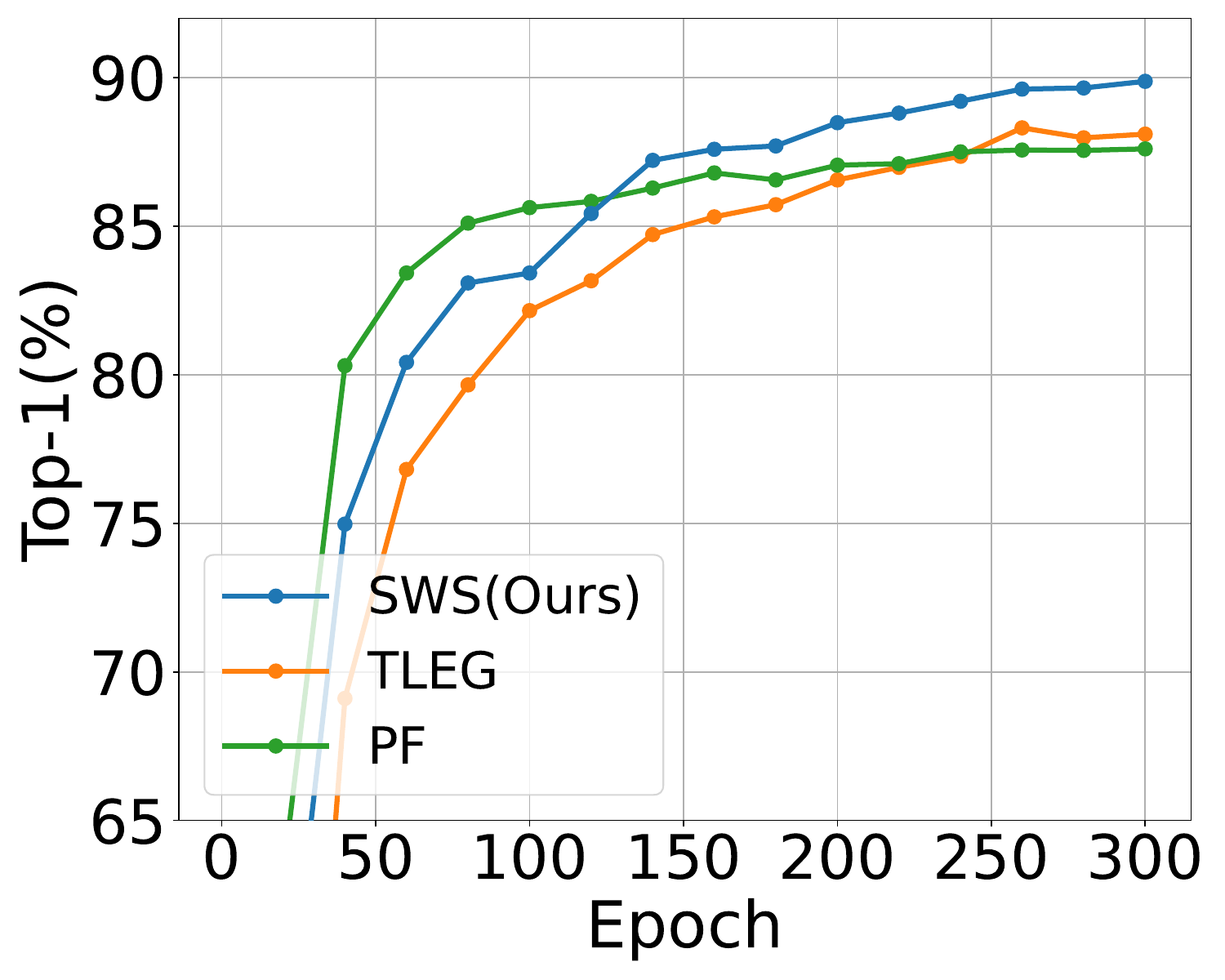}
        % \fbox{\rule{0pt}{1.1in} \rule{.9\linewidth}{0pt}}
        \caption{Cars-196}
        \label{fig6_8}
        \end{subfigure}
        \hfill
        
    \caption{ Performance comparisons on several downstream classification datasets of (a)-(d): Des-S-12 and (e)-(h): Des-B-12.
    }
    % \vspace{-4pt}
    \label{fig:dc_result2}
\end{figure*}

\subsection{Initialization with Learngene}
After obtaining learngene which is composed of well-trained $\{\theta_{1},...,\theta_{M}\}$, we can flexibly initialize Des-Nets of varying depths $L^{ds}$, fitting diverse resource constraints with much less fine-tuning efforts.
In the following, we describe the initialization strategy and the initialization order.

\noindent\textbf{Initialization Strategy.} 
During the learngene learning process, sharing weights in a stage-wise fashion provides important \textit{expansion guidance} for initializing Des-Nets, \textit{i}.\textit{e}., expanding the weight-shared layers at its corresponding stage to initialize the target Des-Net.
Specifically, we present two initialization strategies: \textit{Cyclic Initialization} and \textit{Random-assigned Initialization}.
As shown in Fig.~\ref{fig:method_2}(b), we illustrate the initialization process of a 5-layer Des-Net with 3 learngene layers as an example.
Cyclic Initialization involves the sequential selection of learngene layers in a specific order to initialize corresponding layers of the Des-Net.
In contrast, Random-assigned strategy initializes the Des-Net by randomly selecting learngene layers.
However, as Cyclic Initialization is more aligned with the learngene learning process (\textit{i}.\textit{e}., keeping the stage order the same as the Aux-Net), we will show in Section~\ref{exp_abl} that it achieves more stable and better performance than Random-assigned initialization.
In this case, we take Cyclic Initialization as the default initialization strategy.

\noindent\textbf{Initialization Order.} 
Besides, different initialization orders lead to differences in initialization focus.
For example, in the case of ``Front-Last-Mid'', such initialization focuses more on the front and last part of the Des-Net, as shown in Fig.~\ref{fig:method_2}(c).
With Cyclic Initialization, different initialization orders bring similar performance of Des-Net.  
Therefore, we take ``Front-Mid-Last'' as the default initialization order in SWS.

%%%%%%%%%%%%%%%%%%%%%%%%%%% Experiments 
\section{Experiments}
\subsection{Experimental Setup}
% \label{es}
We perform experiments on ImageNet-1K~\cite{deng2009imagenet} and several downstream datasets including CIFAR-10, CIFAR-100~\cite{krizhevsky2009learning}, Food-101~\cite{bossard2014food} and Cars-196~\cite{krause20133d}.
Model performance is measured by Top-1 classification accuracy (Top-1(\%)).
And we report Params(M) and FLOPs(G) as the number of model parameters and indicators of theoretical complexity of model.
In the first phase, we set two variants of Aux-Net as Aux-S/B where we adopt SWS on DeiT-S/B~\cite{touvron2021training}, after which we train Aux-S/B on ImageNet-1K for 150/100 epochs to obtain learngenes, respectively.
Specifically, we set 5 stages for Aux-S/B, where the shared layer number in each stage is 3,3,4,3 and 3.
We choose Levit-384~\cite{graham2021levit} as the ancestry model.
In the second phase, we set two variants of Des-Net as Des-S/B where we change the layer numbers based on DeiT-S/B, \textit{e}.\textit{g}., we name the 6-layer Des-S as Des-S-6.
Then we initialize Des-S/B with cyclic initialization and fine-tune them for 10 epochs, except that 15 epochs for Des-S-13, Des-S-14 and Des-S-15 for better performance.
% Please see details and numerical results in appendix.
Source code will be available soon.

\begin{table}[t]
	\centering
	\setlength{\tabcolsep}{0.6mm}{
		\begin{tabular}{c|cc|c|cccc}
			\toprule[1.5pt]
			%\cline{2-11}
			\multirow{2}*{$L^{ds}$} & Params & FLOPs & Trained & \multirow{2}*{IMwLM} & Vanilla & \multirow{2}*{TLEG} & \multirow{2}*{SWS} \\
            ~  & (M) & (G) & (100ep) & ~ & -LG & ~ & ~\\
            
            \midrule
            6 & 44.0 & 8.8 & 75.4 & 3.3 & 0.1 & 39.9 & \textbf{59.9}\\
            7 & 51.1 & 10.3 & 76.5 & 5.8 & 0.1 & 60.4 & \textbf{68.5}\\
            8 & 58.2 & 11.7 & 77.2 & 12.2 & 0.1 & 69.8 & \textbf{74.4}\\
            9 & 65.3 & 13.1 & 78.0 & 24.2 & 0.1 & 73.8 & \textbf{76.5}\\
            10 & 72.4 & 14.6 & 78.2 & 42.2 & 0.1 & 75.7 & \textbf{78.0}\\
            11 & 79.5 & 16.0 & 79.0 & 59.2 & 0.1 & 76.4 & \textbf{78.9}\\
            12 & 86.6 & 17.5 & 79.6 & 69.9 & 0.1 & 76.6 & \textbf{79.3}\\
            13 & 93.7 & 18.9 & 79.0 & 76.2 & 0.1 & 76.7 & \textbf{80.0}\\
            14 & 100.7 & 20.4 & 79.1 & 78.6 & 0.1 & 76.5 & \textbf{80.1}\\
            15 & 107.8 & 21.8 & 79.4 & 79.3 & 0.1 & 76.0 & \textbf{80.5}\\

			\bottomrule[1.5pt]
		\end{tabular}
	}
	\caption{Performance comparisons on ImageNet-1K of Des-B with different layer numbers without any tuning after initialization.
    }
    % \vspace{-4pt}
	\label{tab:im-1k_result2}
\end{table}

\begin{table}[t]
	\centering
	\setlength{\tabcolsep}{1.5mm}{
		\begin{tabular}{c|c|cc|cc}
			\toprule[1.5pt]
			%\cline{2-11}
			\multirow{2}*{$L^{ds}$} & Params & \multicolumn{2}{c|}{PF} & \multicolumn{2}{c}{SWS} \\
            ~ & (M) & P-S(M) & Top-1(\%) & P-S(M) & Top-1(\%)\\
            
            \midrule
            6 & 43.4 & 44.0 & 87.99 & \multirow{10}*{37.0} & \textbf{88.11} \\
            7 & 50.4 & 51.1 & 88.04 & ~ & \textbf{88.75} \\
            8 & 57.5 & 58.2 & 88.35 & ~ & \textbf{89.01} \\
            9 & 64.6 & 65.3 & 88.40 & ~ & \textbf{89.35} \\
            10 & 71.7 & 72.4 & 88.04 & ~ & \textbf{89.06} \\
            11 & 78.8 & 79.5 & 88.80 & ~ & \textbf{89.21} \\
            12 & 85.9 & 86.6 & 88.47 & ~ & \textbf{89.59} \\
            13 & 93.0 & 93.7 & 88.36 & ~ & \textbf{89.80} \\
            14 & 100.1 & 100.7 & 88.24 & ~ & \textbf{89.28} \\
            15 & 107.1 & 107.8 & 88.34 & ~ & \textbf{89.20} \\
            
			\bottomrule[1.5pt]
		\end{tabular}
	}
	\caption{Performance comparisons on CIFAR-100 of Des-B with different layer numbers.
    For each target model, PF transfers all the pretrained parameters (P-S(M)) to initialize, which totally requires 759.3M for 10 Des-Bs.
    In contrast, SWS only needs to store 37.0M parameters to initialize each Des-B, which significantly reduces the parameters stored for initialization by 20$\times$ (759.3M \textit{vs}. 37.0M).
    }
    % \vspace{-4pt}
	\label{tab:c100_result1}
\end{table}

\begin{table}[t]
	\centering
	\setlength{\tabcolsep}{2.5mm}{
		\begin{tabular}{c|cc|cc}
			\toprule[1.5pt]
			%\cline{2-11}
			$L^{ds}$ & Params(M) & FLOPs(G)  & Simple-LG & SWS  \\          
            \midrule

            8 & 58.2 & 11.7 & 66.5 & \textbf{74.4} \\
            9 & 65.3 & 13.1 & 63.4 & \textbf{76.5} \\
            10 & 72.4 & 14.6 & 59.9 & \textbf{78.0} \\
            11 & 79.5 & 16.0 & 57.2 & \textbf{78.9} \\
            12 & 86.6 & 17.5 & 52.5  & \textbf{79.3} \\
            
			\bottomrule[1.5pt]
		\end{tabular}
	}
	\caption{Performance comparisons of Des-Bs on ImageNet-1K initialized from learngenes trained with or without SWS.
    Simple-LG means initializing from a normally well-trained 5-layer Des-B.
    }
    % \vspace{-4pt}
	\label{tab:ab_result1}
\end{table}

\begin{table}[t]
	\centering
	\setlength{\tabcolsep}{1.2mm}{
		\begin{tabular}{c|cc|ccc}
			\toprule[1.5pt]
			%\cline{2-11}
			$L^{ds}$ & Params(M) & FLOPs(G) & RI & CI-order1 & CI-order2 \\    
            \midrule

            7 & 51.1 & 10.3 & 50.5 & 68.5 & \textbf{70.0}\\
            8 & 58.2 & 11.7 & 47.8  & \textbf{74.4} & 72.6\\
            9 & 65.3 & 13.1 & 71.2  & \textbf{76.5} & 74.2\\
            10 & 72.4 & 14.6 & 70.7 & \textbf{78.0} & \textbf{78.0}\\
            11 & 79.5 & 16.0 & 76.8 & \textbf{78.9} & \textbf{78.9}\\
            12 & 86.6 & 17.5 & 75.2 & 79.3 & \textbf{79.9}\\
            
			\bottomrule[1.5pt]
		\end{tabular}
	}
	\caption{Performance of Des-Bs on ImageNet-1K under different strategies.
    ``RI'' means Random-assigned Initialization. 
    ``CI-order1'' and ``CI-order2'' mean Cyclic Initialization with different orders.
    }
    % \vspace{-4pt}
	\label{tab:ab_result3}
\end{table}

\begin{table}[t]
	\centering
	\setlength{\tabcolsep}{1.5mm}{
		\begin{tabular}{c|cc|c}
			\toprule[1.5pt]
			%\cline{2-11}
			\multirow{2}*{Method} & Params & FLOPs & Top-1\\     
            ~ & (M) & (G) & (\%) \\
            
            \midrule
            Scratch & 86.6 & 17.5 & 79.6\\
            TLEG~\cite{xia2024transformer} & 15.7 & 17.6 & 76.7\\
            Mini-DeiT~\cite{zhang2022minivit} & 44.1 & 17.5 & \textbf{80.9}\\
            
            \midrule
            MS-WS1 (2,2,2,2,2,2) & 44.0 & 17.5 & 80.7\\
            MS-WS1 (wo dis) & 44.0 & 17.5 & 77.7\\
            MS-WS2 (1,4,1,1,4,1) & 44.0 & 17.5 & 80.0\\
            MS-WS3 (1,5,1,4,1) & 37.0 & 17.5 & 79.4\\
            MS-WS4 (4,1,6,1,4) & 37.0 & 23.3 & 80.7\\
            MS-WS5 (4,1,5,2,3) & 37.0 & 21.8 & fail\\
            MS-WS6 (3,3,4,3,3) & 37.0 & 23.3 & \textbf{80.9}\\
            
			\bottomrule[1.5pt]
		\end{tabular}
	}
    \caption{Performance of Aux-B on ImageNet-1K under different SWS strategies.
    Number in brackets means the shared layer number in each stage, separated by commas. 
    ``wo dis'' means training without distillation.
    Take ``MS-WS6 (3,3,4,3,3)'' as an example, it means 5 stages where the shared layer number in each stage is 3,3,4,3,3.
    }
    % \vspace{-10pt}
	\label{tab:ab_result2}
\end{table}

\begin{table}[t]
	\centering
	\setlength{\tabcolsep}{1.3mm}{
		\begin{tabular}{c|cccc}
			\toprule[1.5pt]
			%\cline{2-11}
			Aux-S & CIFAR-100 & CIFAR-10 & Food-101 & Cars-196 \\    
            \midrule

            76.4 (100) & 87.40 & 98.26 & 89.19 & 84.50 \\
            78.7 (150) & \textbf{88.54} & \textbf{98.58} & \textbf{89.83} & \textbf{87.88} \\
            
			\bottomrule[1.5pt]
		\end{tabular}
	}
	\caption{Performance of Des-S-12 with learngenes trained under different epochs.
    ``Aux-S'' means the performance of Aux-S on ImageNet-1K.
    The number in bracket means training epochs. 
    }
    % \vspace{-10pt}
	\label{tab:ab_result4}
\end{table}

\subsection{Main Results}
\noindent\textbf{SWS achieves better performance while reducing huge training efforts in contrast to from scratch training on ImageNet-1K.}
We report ImageNet-1K classification performance of 20 different Des-Nets in Fig.~\ref{fig:im1k_result1}, where ``Scratch'' denotes training from scratch, ``TLEG'' denotes linearly expanding learngenes to initialize~\cite{xia2024transformer}.
Compared to Scratch, SWS can achieve better performance and significantly improve training efficiency.
Take 10 Des-Bs as an example, SWS performs better while reducing around \textbf{6.6$\times$} total training costs (10$\times$100 epochs \textit{vs}. 100+10$\times$5 epochs), compared to training each Des-B from scratch for 100 epochs.
For each Des-B, SWS can reduce around \textbf{20$\times$} training costs.
In some cases such as from Des-B-10 to Des-B-15, SWS performs better only after \textbf{1 epoch} tuning, which demonstrates the effectiveness of initialization via SWS.
Compared to TLEG, SWS performs better and further enhances the efficiency.
Take 10 Des-Bs as an example, SWS performs better while reducing around \textbf{2.5$\times$} total training costs (100+10$\times$40 epochs \textit{vs}. 100+10$\times$10 epochs).
In a nutshell, the efficiency of SWS becomes more obvious with the number of Des-Nets increasing as we only need to train learngenes \emph{once}.

\noindent\textbf{When transferring to downstream classification datasets, SWS presents competitive results.}
We compare SWS against pre-training and fine-tuning (\textbf{PF}), Scratch, Vanilla-LG~\cite{wang2022learngene} and TLEG~\cite{xia2024transformer} on 4 classification datasets.
As shown in Fig.~\ref{fig:dc_result2}, we observe that SWS consistently outperforms several baselines, which verifies the effectiveness of initializing with learngenes trained via SWS.
Take Des-B-12 as an example, SWS consistently outperforms PF by \boldmath{$1.12\%$, $0.35\%$, $1.65\%$ and $2.28\%$} respectively on CIFAR-100, CIFAR-10, Food-101 and Cars-196.

\noindent\textbf{SWS significantly outperforms initialization baselines when directly evaluating on ImageNet-1K without any tuning after initialization.}
To validate the initialization quality of SWS, we compare SWS against Trained, Vanilla-LG~\cite{wang2022learngene}, TLEG~\cite{xia2024transformer} and IMwLM~\cite{xu2023initializing} on ImageNet-1K, where Trained means models trained from scratch for 100 epochs and IMwLM means initializing small models from a larger model.
From Table~\ref{tab:im-1k_result2}, we observe that SWS significantly outperforms all baselines by a large margin.
For example, SWS outperforms IMwLM by \boldmath{$52.3\%$, $35.8\%$, $19.7\%$ and $9.4\%$} respectively on Des-B-9, Des-B-10, Des-B-11 and Des-B-12.
Notably, we also find that directly initializing via SWS \textit{without any tuning} can achieve comparable performance with well-trained models.
For example, SWS outperforms Trained by \boldmath{$1.0\%$, $1.0\%$ and $1.1\%$} respectively on Des-B-13, Des-B-14 and Des-B-15, which verifies that the stage-wise information has been preserved to learngenes.

\noindent\textbf{When initializing variable-sized models, SWS notably reduces the parameters stored to initialize compared with pre-training and fine-tuning (PF).}
We report the results of Des-Bs initialized from learngenes and those initialized from pretrained parameters whose number equals to that of target model.
From Table~\ref{tab:c100_result1}, SWS achieves better performance and significantly reduces \textbf{20$\times$} (759.3M \textit{vs}. 37.0M) parameters stored to initialize, in constrast to PF.
Furthermore, SWS only needs to train learngene \textit{once} while PF requires pretraining each Des-B individually, thereby significantly reducing the pre-training costs.
Specifically, SWS reduces pre-training costs by \textbf{10$\times$} (10$\times$100 epochs \textit{vs}. 1$\times$100 epochs) compared to PF.
It is noteworthy that the efficiency of SWS becomes more obvious with the increasing number of Des-Nets.

\subsection{Ablation and Analysis}
\label{exp_abl}
We investigate from (1) training learngenes with or without stage-wise weight sharing, (2) training learngenes under different stage-wise weight sharing strategies, (3) initializing Des-Nets under different strategies, (4) initializing Des-Nets with learngenes trained under different epochs.

\noindent\textbf{The effect of stage-wise weight sharing.}
As shown in Table~\ref{tab:ab_result1}, we observe that SWS significantly outperforms Simple-LG when directly evaluating on ImageNet-1K without any tuning after initialization, which demonstrates the importance of stage information and expansion guidance for initializing Des-Nets.
For example, SWS outperforms Simple-LG by \boldmath{$18.1\%$ and $21.7\%$} respectively on Des-B-10 and Des-B-11.

\noindent\textbf{The effect of different stage-wise weight sharing strategies.}
We present the performance of Aux-B to show the quality of trained learngenes.
From Table~\ref{tab:ab_result2}, we observe that MS-WS6 outperforms MS-WS4 and MS-WS5 fails to converge, which reflects balanced sharing is more stable and better than unbalanced one.
Moreover, we find that MS-WS4 outperforms MS-WS3, which demonstrates starting from a weight-sharing stage is better.
Also, MS-WS5 achieves comparable performance in contrast to Mini-DeiT~\cite{zhang2022minivit}.

\noindent\textbf{The effect of different initialization strategies.}
From Table~\ref{tab:ab_result3}, we find that performance of Des-Nets with Cyclic Initialization (CI) outperforms that with Random-assigned Initialization (RI) when directly evaluating on ImageNet-1K without any tuning after initialization.
For example, CI-order1 outperforms RI by \boldmath{$5.3\%$ and $7.3\%$} respectively on Des-B-9 and Des-B-10.

\noindent\textbf{The effect of learngenes on Des-Nets.}
We train Aux-S for more epochs.
As shown in Table~\ref{tab:ab_result4}, we find that better performance of Des-S-12 can be consistently achieved with better learngenes.
For example, performance on CIFAR-100 can be improved from $87.40\%$ to $88.54\%$ with better learngenes.

%%%%%%%%%%%%%%%%%%%%%%%%%%% Conclusion 
\section{Conclusion}
\label{sec:conclusion}
In this paper, we proposed a well-motivated and highly effective Learngene approach termed SWS to initialize variable-sized Transformers, enabling adaptation to diverse resource constraints.
Experimental results under various initialization settings demonstrated the effectiveness and efficiency of SWS.

%% The file named.bst is a bibliography style file for BibTeX 0.99c
\bibliographystyle{named}
\bibliography{ijcai24}

\begin{thebibliography}{}

\bibitem[\protect\citeauthoryear{Arpit \bgroup \em et al.\egroup
  }{2019}]{arpit2019initialize}
Devansh Arpit, V{\'\i}ctor Campos, and Yoshua Bengio.
\newblock How to initialize your network? robust initialization for weightnorm
  \& resnets.
\newblock {\em Advances in Neural Information Processing Systems}, 32, 2019.

\bibitem[\protect\citeauthoryear{Ba \bgroup \em et al.\egroup
  }{2016}]{ba2016layer}
Jimmy~Lei Ba, Jamie~Ryan Kiros, and Geoffrey~E Hinton.
\newblock Layer normalization.
\newblock {\em arXiv preprint arXiv:1607.06450}, 2016.

\bibitem[\protect\citeauthoryear{Bai \bgroup \em et al.\egroup
  }{2019}]{bai2019deep}
Shaojie Bai, J~Zico Kolter, and Vladlen Koltun.
\newblock Deep equilibrium models.
\newblock {\em Advances in Neural Information Processing Systems}, 32, 2019.

\bibitem[\protect\citeauthoryear{Bao \bgroup \em et al.\egroup
  }{2022}]{bao2021beit}
Hangbo Bao, Li~Dong, Songhao Piao, and Furu Wei.
\newblock Beit: Bert pre-training of image transformers.
\newblock {\em ICLR}, 2022.

\bibitem[\protect\citeauthoryear{Bossard \bgroup \em et al.\egroup
  }{2014}]{bossard2014food}
Lukas Bossard, Matthieu Guillaumin, and Luc Van~Gool.
\newblock Food-101--mining discriminative components with random forests.
\newblock In {\em Computer Vision--ECCV 2014: 13th European Conference, Zurich,
  Switzerland, September 6-12, 2014, Proceedings, Part VI 13}, pages 446--461.
  Springer, 2014.

\bibitem[\protect\citeauthoryear{Dabre and Fujita}{2019}]{dabre2019recurrent}
Raj Dabre and Atsushi Fujita.
\newblock Recurrent stacking of layers for compact neural machine translation
  models.
\newblock In {\em Proceedings of the AAAI Conference on Artificial
  Intelligence}, volume~33, pages 6292--6299, 2019.

\bibitem[\protect\citeauthoryear{Deng \bgroup \em et al.\egroup
  }{2009}]{deng2009imagenet}
Jia Deng, Wei Dong, Richard Socher, Li-Jia Li, Kai Li, and Li~Fei-Fei.
\newblock Imagenet: A large-scale hierarchical image database.
\newblock In {\em 2009 IEEE conference on computer vision and pattern
  recognition}, pages 248--255. Ieee, 2009.

\bibitem[\protect\citeauthoryear{Devlin \bgroup \em et al.\egroup
  }{2018}]{devlin2018bert}
Jacob Devlin, Ming-Wei Chang, Kenton Lee, and Kristina Toutanova.
\newblock Bert: Pre-training of deep bidirectional transformers for language
  understanding.
\newblock {\em arXiv preprint arXiv:1810.04805}, 2018.

\bibitem[\protect\citeauthoryear{Dosovitskiy \bgroup \em et al.\egroup
  }{2021}]{dosovitskiy2020image}
Alexey Dosovitskiy, Lucas Beyer, Alexander Kolesnikov, Dirk Weissenborn,
  Xiaohua Zhai, Thomas Unterthiner, Mostafa Dehghani, Matthias Minderer, Georg
  Heigold, Sylvain Gelly, et~al.
\newblock An image is worth 16x16 words: Transformers for image recognition at
  scale.
\newblock {\em ICLR}, 2021.

\bibitem[\protect\citeauthoryear{Feng \bgroup \em et al.\egroup
  }{2024}]{feng2024transferring}
Fu~Feng, Jing Wang, and Xin Geng.
\newblock Transferring core knowledge via learngenes.
\newblock {\em arXiv preprint arXiv:2401.08139}, 2024.

\bibitem[\protect\citeauthoryear{Glorot and
  Bengio}{2010}]{glorot2010understanding}
Xavier Glorot and Yoshua Bengio.
\newblock Understanding the difficulty of training deep feedforward neural
  networks.
\newblock In {\em Proceedings of the thirteenth international conference on
  artificial intelligence and statistics}, pages 249--256. JMLR Workshop and
  Conference Proceedings, 2010.

\bibitem[\protect\citeauthoryear{Graham \bgroup \em et al.\egroup
  }{2021}]{graham2021levit}
Benjamin Graham, Alaaeldin El-Nouby, Hugo Touvron, Pierre Stock, Armand Joulin,
  Herv{\'e} J{\'e}gou, and Matthijs Douze.
\newblock Levit: a vision transformer in convnet's clothing for faster
  inference.
\newblock In {\em Proceedings of the IEEE/CVF international conference on
  computer vision}, pages 12259--12269, 2021.

\bibitem[\protect\citeauthoryear{He \bgroup \em et al.\egroup
  }{2015}]{he2015delving}
Kaiming He, Xiangyu Zhang, Shaoqing Ren, and Jian Sun.
\newblock Delving deep into rectifiers: Surpassing human-level performance on
  imagenet classification.
\newblock In {\em Proceedings of the IEEE international conference on computer
  vision}, pages 1026--1034, 2015.

\bibitem[\protect\citeauthoryear{He \bgroup \em et al.\egroup
  }{2022}]{he2022masked}
Kaiming He, Xinlei Chen, Saining Xie, Yanghao Li, Piotr Doll{\'a}r, and Ross
  Girshick.
\newblock Masked autoencoders are scalable vision learners.
\newblock In {\em Proceedings of the IEEE/CVF conference on computer vision and
  pattern recognition}, pages 16000--16009, 2022.

\bibitem[\protect\citeauthoryear{Hinton \bgroup \em et al.\egroup
  }{2015}]{hinton2015distilling}
Geoffrey Hinton, Oriol Vinyals, and Jeff Dean.
\newblock Distilling the knowledge in a neural network.
\newblock {\em arXiv preprint arXiv:1503.02531}, 2015.

\bibitem[\protect\citeauthoryear{Huang \bgroup \em et al.\egroup
  }{2020}]{huang2020improving}
Xiao~Shi Huang, Felipe Perez, Jimmy Ba, and Maksims Volkovs.
\newblock Improving transformer optimization through better initialization.
\newblock In {\em International Conference on Machine Learning}, pages
  4475--4483. PMLR, 2020.

\bibitem[\protect\citeauthoryear{Ji \bgroup \em et al.\egroup
  }{2023}]{ji2023teachers}
Zhong Ji, Jingwei Ni, Xiyao Liu, and Yanwei Pang.
\newblock Teachers cooperation: team-knowledge distillation for multiple
  cross-domain few-shot learning.
\newblock {\em Frontiers of Computer Science}, 17(2):172312, 2023.

\bibitem[\protect\citeauthoryear{Kovaleva \bgroup \em et al.\egroup
  }{2019}]{kovaleva2019revealing}
Olga Kovaleva, Alexey Romanov, Anna Rogers, and Anna Rumshisky.
\newblock Revealing the dark secrets of bert.
\newblock {\em EMNLP}, 2019.

\bibitem[\protect\citeauthoryear{Krause \bgroup \em et al.\egroup
  }{2013}]{krause20133d}
Jonathan Krause, Michael Stark, Jia Deng, and Li~Fei-Fei.
\newblock 3d object representations for fine-grained categorization.
\newblock In {\em Proceedings of the IEEE international conference on computer
  vision workshops}, pages 554--561, 2013.

\bibitem[\protect\citeauthoryear{Krizhevsky \bgroup \em et al.\egroup
  }{2009}]{krizhevsky2009learning}
Alex Krizhevsky, Geoffrey Hinton, et~al.
\newblock Learning multiple layers of features from tiny images.
\newblock {\em Technique Report}, 2009.

\bibitem[\protect\citeauthoryear{Lan \bgroup \em et al.\egroup
  }{2020}]{lan2019albert}
Zhenzhong Lan, Mingda Chen, Sebastian Goodman, Kevin Gimpel, Piyush Sharma, and
  Radu Soricut.
\newblock Albert: A lite bert for self-supervised learning of language
  representations.
\newblock {\em ICLR}, 2020.

\bibitem[\protect\citeauthoryear{LeCun \bgroup \em et al.\egroup
  }{2002}]{lecun2002efficient}
Yann LeCun, L{\'e}on Bottou, Genevieve~B Orr, and Klaus-Robert M{\"u}ller.
\newblock Efficient backprop.
\newblock In {\em Neural networks: Tricks of the trade}, pages 9--50. Springer,
  2002.

\bibitem[\protect\citeauthoryear{Li \bgroup \em et al.\egroup
  }{2024}]{li2024accelerating}
Lei Li, Chengyu Wang, Minghui Qiu, Cen Chen, Ming Gao, and Aoying Zhou.
\newblock Accelerating bert inference with gpu-efficient exit prediction.
\newblock {\em Frontiers of Computer Science}, 18(3):183308, 2024.

\bibitem[\protect\citeauthoryear{Li \bgroup \em et al.\egroup
  }{2025}]{li2025kdcrowd}
Shaoyuan Li, Yuxiang Zheng, Ye~Shi, Shengjun Huang, and Songcan Chen.
\newblock Kd-crowd: A knowledge distillation framework for learning from
  crowds.
\newblock {\em Frontiers of Computer Science}, 19(1):191302, 2025.

\bibitem[\protect\citeauthoryear{Liu \bgroup \em et al.\egroup
  }{2021}]{liu2021swin}
Ze~Liu, Yutong Lin, Yue Cao, Han Hu, Yixuan Wei, Zheng Zhang, Stephen Lin, and
  Baining Guo.
\newblock Swin transformer: Hierarchical vision transformer using shifted
  windows.
\newblock In {\em Proceedings of the IEEE/CVF international conference on
  computer vision}, pages 10012--10022, 2021.

\bibitem[\protect\citeauthoryear{Mishkin and Matas}{2015}]{mishkin2015all}
Dmytro Mishkin and Jiri Matas.
\newblock All you need is a good init.
\newblock {\em arXiv preprint arXiv:1511.06422}, 2015.

\bibitem[\protect\citeauthoryear{Nair and Hinton}{2010}]{nair2010rectified}
Vinod Nair and Geoffrey~E Hinton.
\newblock Rectified linear units improve restricted boltzmann machines.
\newblock In {\em Proceedings of the 27th international conference on machine
  learning (ICML-10)}, pages 807--814, 2010.

\bibitem[\protect\citeauthoryear{Oquab \bgroup \em et al.\egroup
  }{2023}]{oquab2023dinov2}
Maxime Oquab, Timoth{\'e}e Darcet, Th{\'e}o Moutakanni, Huy Vo, Marc
  Szafraniec, Vasil Khalidov, Pierre Fernandez, Daniel Haziza, Francisco Massa,
  Alaaeldin El-Nouby, et~al.
\newblock Dinov2: Learning robust visual features without supervision.
\newblock {\em arXiv preprint arXiv:2304.07193}, 2023.

\bibitem[\protect\citeauthoryear{Paszke \bgroup \em et al.\egroup
  }{2019}]{paszke2019pytorch}
Adam Paszke, Sam Gross, Francisco Massa, Adam Lerer, James Bradbury, Gregory
  Chanan, Trevor Killeen, Zeming Lin, Natalia Gimelshein, Luca Antiga, et~al.
\newblock Pytorch: An imperative style, high-performance deep learning library.
\newblock In {\em Proceedings of the 33th Annual Conference on Neural
  Information Processing Systems}, pages 8024--8035, 2019.

\bibitem[\protect\citeauthoryear{Radford \bgroup \em et al.\egroup
  }{2021}]{radford2021learning}
Alec Radford, Jong~Wook Kim, Chris Hallacy, Aditya Ramesh, Gabriel Goh,
  Sandhini Agarwal, Girish Sastry, Amanda Askell, Pamela Mishkin, Jack Clark,
  et~al.
\newblock Learning transferable visual models from natural language
  supervision.
\newblock In {\em International conference on machine learning}, pages
  8748--8763. PMLR, 2021.

\bibitem[\protect\citeauthoryear{Ren \bgroup \em et al.\egroup
  }{2023}]{ren2023tinymim}
Sucheng Ren, Fangyun Wei, Zheng Zhang, and Han Hu.
\newblock Tinymim: An empirical study of distilling mim pre-trained models.
\newblock In {\em Proceedings of the IEEE/CVF Conference on Computer Vision and
  Pattern Recognition}, pages 3687--3697, 2023.

\bibitem[\protect\citeauthoryear{Samragh \bgroup \em et al.\egroup
  }{2023}]{samragh2023weight}
Mohammad Samragh, Mehrdad Farajtabar, Sachin Mehta, Raviteja Vemulapalli,
  Fartash Faghri, Devang Naik, Oncel Tuzel, and Mohammad Rastegari.
\newblock Weight subcloning: direct initialization of transformers using larger
  pretrained ones.
\newblock {\em arXiv preprint arXiv:2312.09299}, 2023.

\bibitem[\protect\citeauthoryear{Shi \bgroup \em et al.\egroup
  }{2024}]{shi2024building}
Boyu Shi, Shiyu Xia, Xu~Yang, Haokun Chen, Zhiqiang Kou, and Xin Geng.
\newblock Building variable-sized models via learngene pool.
\newblock In {\em Proceedings of the AAAI Conference on Artificial
  Intelligence}, volume~38, pages 14946--14954, 2024.

\bibitem[\protect\citeauthoryear{Takase and Kiyono}{2021}]{takase2021lessons}
Sho Takase and Shun Kiyono.
\newblock Lessons on parameter sharing across layers in transformers.
\newblock {\em arXiv preprint arXiv:2104.06022}, 2021.

\bibitem[\protect\citeauthoryear{Touvron \bgroup \em et al.\egroup
  }{2021}]{touvron2021training}
Hugo Touvron, Matthieu Cord, Matthijs Douze, Francisco Massa, Alexandre
  Sablayrolles, and Herv{\'e} J{\'e}gou.
\newblock Training data-efficient image transformers \& distillation through
  attention.
\newblock In {\em International conference on machine learning}, pages
  10347--10357. PMLR, 2021.

\bibitem[\protect\citeauthoryear{Touvron \bgroup \em et al.\egroup
  }{2023}]{touvron2023llama}
Hugo Touvron, Thibaut Lavril, Gautier Izacard, Xavier Martinet, Marie-Anne
  Lachaux, Timoth{\'e}e Lacroix, Baptiste Rozi{\`e}re, Naman Goyal, Eric
  Hambro, Faisal Azhar, et~al.
\newblock Llama: Open and efficient foundation language models.
\newblock {\em arXiv preprint arXiv:2302.13971}, 2023.

\bibitem[\protect\citeauthoryear{Wang \bgroup \em et al.\egroup
  }{2022a}]{wang2022learngene}
Qiu-Feng Wang, Xin Geng, Shu-Xia Lin, Shi-Yu Xia, Lei Qi, and Ning Xu.
\newblock Learngene: From open-world to your learning task.
\newblock In {\em Proceedings of the AAAI Conference on Artificial
  Intelligence}, volume~36, pages 8557--8565, 2022.

\bibitem[\protect\citeauthoryear{Wang \bgroup \em et al.\egroup
  }{2022b}]{wang2022image}
Wenhui Wang, Hangbo Bao, Li~Dong, Johan Bjorck, Zhiliang Peng, Qiang Liu, Kriti
  Aggarwal, Owais~Khan Mohammed, Saksham Singhal, Subhojit Som, et~al.
\newblock Image as a foreign language: Beit pretraining for all vision and
  vision-language tasks.
\newblock {\em arXiv preprint arXiv:2208.10442}, 2022.

\bibitem[\protect\citeauthoryear{Wang \bgroup \em et al.\egroup
  }{2023}]{wang2023learngene}
Qiufeng Wang, Xu~Yang, Shuxia Lin, and Xin Geng.
\newblock Learngene: Inheriting condensed knowledge from the ancestry model to
  descendant models.
\newblock {\em arXiv preprint arXiv:2305.02279}, 2023.

\bibitem[\protect\citeauthoryear{Xia \bgroup \em et al.\egroup
  }{2024}]{xia2024transformer}
Shiyu Xia, Miaosen Zhang, Xu~Yang, Ruiming Chen, Haokun Chen, and Xin Geng.
\newblock Transformer as linear expansion of learngene.
\newblock In {\em Proceedings of the AAAI Conference on Artificial
  Intelligence}, volume~38, pages 16014--16022, 2024.

\bibitem[\protect\citeauthoryear{Xu \bgroup \em et al.\egroup
  }{2023}]{xu2023initializing}
Zhiqiu Xu, Yanjie Chen, Kirill Vishniakov, Yida Yin, Zhiqiang Shen, Trevor
  Darrell, Lingjie Liu, and Zhuang Liu.
\newblock Initializing models with larger ones.
\newblock {\em arXiv preprint arXiv:2311.18823}, 2023.

\bibitem[\protect\citeauthoryear{Zhang \bgroup \em et al.\egroup
  }{2022}]{zhang2022minivit}
Jinnian Zhang, Houwen Peng, Kan Wu, Mengchen Liu, Bin Xiao, Jianlong Fu, and
  Lu~Yuan.
\newblock Minivit: Compressing vision transformers with weight multiplexing.
\newblock In {\em Proceedings of the IEEE/CVF Conference on Computer Vision and
  Pattern Recognition}, pages 12145--12154, 2022.

\end{thebibliography}

\end{document}